\documentclass[conference]{IEEEtran}
\IEEEoverridecommandlockouts
\usepackage{cite}
\usepackage{amsmath,amssymb,amsfonts}
\usepackage{algorithm}
\usepackage{algorithmic}
\usepackage{graphicx}
\usepackage{textcomp}
\usepackage{xcolor}
\usepackage{subfigure}
\usepackage{hyperref}

\def\BibTeX{{\rm B\kern-.05em{\sc i\kern-.025em b}\kern-.08em
    T\kern-.1667em\lower.7ex\hbox{E}\kern-.125emX}}
\begin{document}

\title{Simplifying Reinforced Feature Selection via Restructured Choice Strategy of Single Agent}

\author{\IEEEauthorblockN{Xiaosa Zhao\IEEEauthorrefmark{1}, Kunpeng Liu\IEEEauthorrefmark{2}, Wei Fan\IEEEauthorrefmark{2}, Lu Jiang\IEEEauthorrefmark{1},  Xiaowei Zhao\IEEEauthorrefmark{1}, Minghao Yin\IEEEauthorrefmark{1}, Yanjie Fu\IEEEauthorrefmark{2}}
\IEEEauthorblockA{\textit{\IEEEauthorrefmark{1} Northeast Normal University, Changchun} \\
\textit{\IEEEauthorrefmark{2} Department of Computer Science, University of Central Florida, Orlando} \\
\IEEEauthorrefmark{1}\{zhaoxs686, jiangl761, zhaoxw303, ymh \}@nenu.edu.cn, \\ 
\IEEEauthorrefmark{2}\{kunpengliu, weifan\}@knights.ucf.edu, 
\IEEEauthorrefmark{2}\{Yanjie.Fu\}@ucf.edu \\
}
\IEEEcompsocitemizethanks{
	\IEEEcompsocthanksitem \textbullet ~ Kunpeng Liu is co-first author.
	\IEEEcompsocthanksitem \textbullet ~ Yanjie Fu is corresponding author.
	}

}

\maketitle

\begin{abstract}

Feature selection aims to select a subset of features to optimize the performances of downstream predictive tasks. 
Recently, multi-agent reinforced feature selection (MARFS) has been introduced to automate feature selection, by creating agents for each feature to select or deselect corresponding features. 
Although MARFS enjoys the automation of the selection process, MARFS suffers from not just the data complexity in terms of contents and dimensionality, but also the exponentially-increasing computational costs with regard to the number of agents.
The raised concern leads to a new research question: 
Can we simplify the selection process of agents under reinforcement learning context so as to improve the efficiency and costs of feature selection?
To address the question, we develop a single-agent reinforced feature selection approach integrated with restructured choice strategy. 
Specifically, the restructured choice strategy includes: 
1) we exploit only one single agent to handle the selection task of multiple features, instead of using multiple agents. 
2) we develop a scanning method to empower the single agent to make multiple selection/deselection decisions in each round of scanning. 
3) we exploit the relevance to predictive labels of features to prioritize the scanning orders of the agent for multiple features. 
4) we propose a convolutional auto-encoder algorithm, integrated with the encoded index information of features, to improve state representation. 
5) we design a reward scheme that take into account both prediction accuracy and feature redundancy to facilitate the exploration process.
Finally, we present extensive experimental results to demonstrate the efficiency and effectiveness of the proposed method.

\end{abstract}


\section{Introduction}
Feature selection is a classic and one of the most important problems in machine learning and data mining field, aiming to remove irrelevant and redundant features from the original feature set. With effective and efficient feature selection algorithms, we can achieve low computation cost, good performance of the machine learning pipeline as well as good interpretation of the learning process.

Traditional feature selection methods include filter methods, wrapper methods and embedded methods \cite{b4}. Filter methods evaluate features based on their statistical characteristics, e.g., Fisher score \cite{b5}, correlation coefficient \cite{b6} and information gain (IG) \cite{b7}. Filter methods are independent of the downstream machine learning model, and the representative methods are maximum relevance minimum redundancy (mRMR) \cite{b8}, fast correlation based filter algorithm (FCBF)\cite{yu2004efficient}, and univariate feature selection \cite{forman2003extensive}. Wrapper methods evaluate the quality of the selected features with a predetermined classifier, and the representative methods are recursive feature elimination (RFE) \cite{b9} \cite{guyon2002gene}, sequential forward selection (SFS) \cite{b10}, and evolutionary algorithms \cite{b11, b12, b13, b14}. Embedded methods incorporate feature selection as a part of classification model in the training phase, and the typical methods are LARS \cite{efron2004least}, LASSO \cite{tibshirani1996regression}, and decision tree \cite{sugumaran2007feature}. These methods usually suffer from locality and can not achieve very good performance.

Recently, reinforcement learning (RL) methods are introduced to the field of feature selection. Model-free reinforcement learning can obtain long-term optimal decisions in an unknown environment by interacting with dynamic environment, keeping the exploration-exploitation trade-off, and receiving a reward signal as feedback \cite{b16}. These advantages enable reinforcement learning the ability of globally automating feature subspace exploration and thus can achieve better performance than traditional methods. \emph{Liu et al.} proposed a multi-agent reinforcement learning framework (MARLFS) for feature selection problem by regarding each feature as an agent, where the agent can exchange information in the dynamically selected feature subspace \cite{b19}. Sharing experience can help agent to capture the interaction between features and greatly benefits the feature subspace exploration process. However, the complexity of the MARLFS increases with the number of agents, as each agent has to maintain its own policy network, training strategy and memory storage \cite{b20}. When the feature number in the dataset is large, this algorithm puts much pressure to the computational platform. It is naturally promising to investigate if we can solve the feature selection problem with a single-agent framework which essentially requires less computational resources and meanwhile can still maintain the benefits of the reinforcement learning framework. However, several challenges arise.

The first challenge is to formulate the feature selection problem with the single-agent reinforcement learning framework. Intuitively, we can design a single agent which determines the selection and deselection once for all the $N$ features \cite{b17} \cite{b18}. However, the action space of these methods are $2^N$, making the feature selection NP-hard and can only achieve local optima. In this paper, we propose a scanning strategy, where the agent scans the features one by one to decide the selection. After scanning one feature, we derive the state representation as well as the reward scheme, which comprises prediction accuracy and feature redundancy. With new policy network, the agent goes to the next feature to decide its selection/deselection. After scanning all the features, we have a selected feature subset. With more and more scanning episodes, the policy network will be better and better and thus can select more reasonable feature subsets.

The second challenge is to decide the scanning order. Different scanning order of the features may bring on different selected feature subset. For example, suppose we have the feature space including $\{f_1, f_2, f_3\}$. If the scanning order is $f_1 \rightarrow f_2 \rightarrow f_3$, the actions could be 'selection', 'deselection' and 'selection', and the final selected feature subset is $\{f_1, f_3\}$. However, if the scanning order is $f_3 \rightarrow f_1 \rightarrow f_2$, the actions could be 'deselection','selection' and 'selection', and the final selected feature subset is $\{f_2$,$f_3\}$. This is because with different scanning order, the agent will receive different reward and thus get its policy network trained to different directions. In this paper, to achieve an optimal scanning order, we propose a relevance-based scanning strategy. To be more specific, we calculate the relevance of each feature with the label. The higher the relevance is, the earlier the related feature will be scanned. 

The third challenge is to accurately represent the environment. We need to derive the state representation from the selected feature subset, which is dynamically changing in the feature selection process. However, the policy network of reinforcement learning requires a fixed-length state representation vector. To solve this problem, existing studies proposed three methods, i.e., meta descriptive statistics, auto-encoder based deep representation and dynamic-graph based graph convolutional network (GCN) \cite{b19}. The limitation of Meta descriptive statistics method is unsatisfactory performance, while the limitation of GCN method is the strong assumption of full-connected graph. In this paper, we further extend the auto-encoder based method with convolutional neural network and propose a convolutional auto-encoder representation method. In order to solve the multi-size input in the encoder layer, we replace the pooling layer after the last convolutional layer with the spatial pyramid pooling layer \cite{b23}. So, in this convolutional auto-encoder, the encoder layer contains convolutional layers, pooling layers and a spatial pyramid pooling layer, and then the decoder layer contains an inverse spatial pyramid pooling layer, unsampling layers and convolutional layers. In addition, to identify the feature which the agent is scanning, we incorporate its index information into the state representation.

To summarize, in this paper, we propose a scanning based single-agent reinforcement learning feature selection method. Our major contributions are: (1) we formulate the feature selection problem by incorporating a scanning strategy into the single-agent reinforcement learning framework, where the agent scans and selects features one by one (2) we propose a relevance-based scanning strategy which regulates the scanning order to improve the learning performance; (3) we propose a convolutional auto-encoder representation method to derive accurate state vector; (4) we design comprehensive experiments to show the superiority of the propose method.

\begin{figure*}[t]
\vspace{0.1cm}
\centerline{\includegraphics[width=17cm]{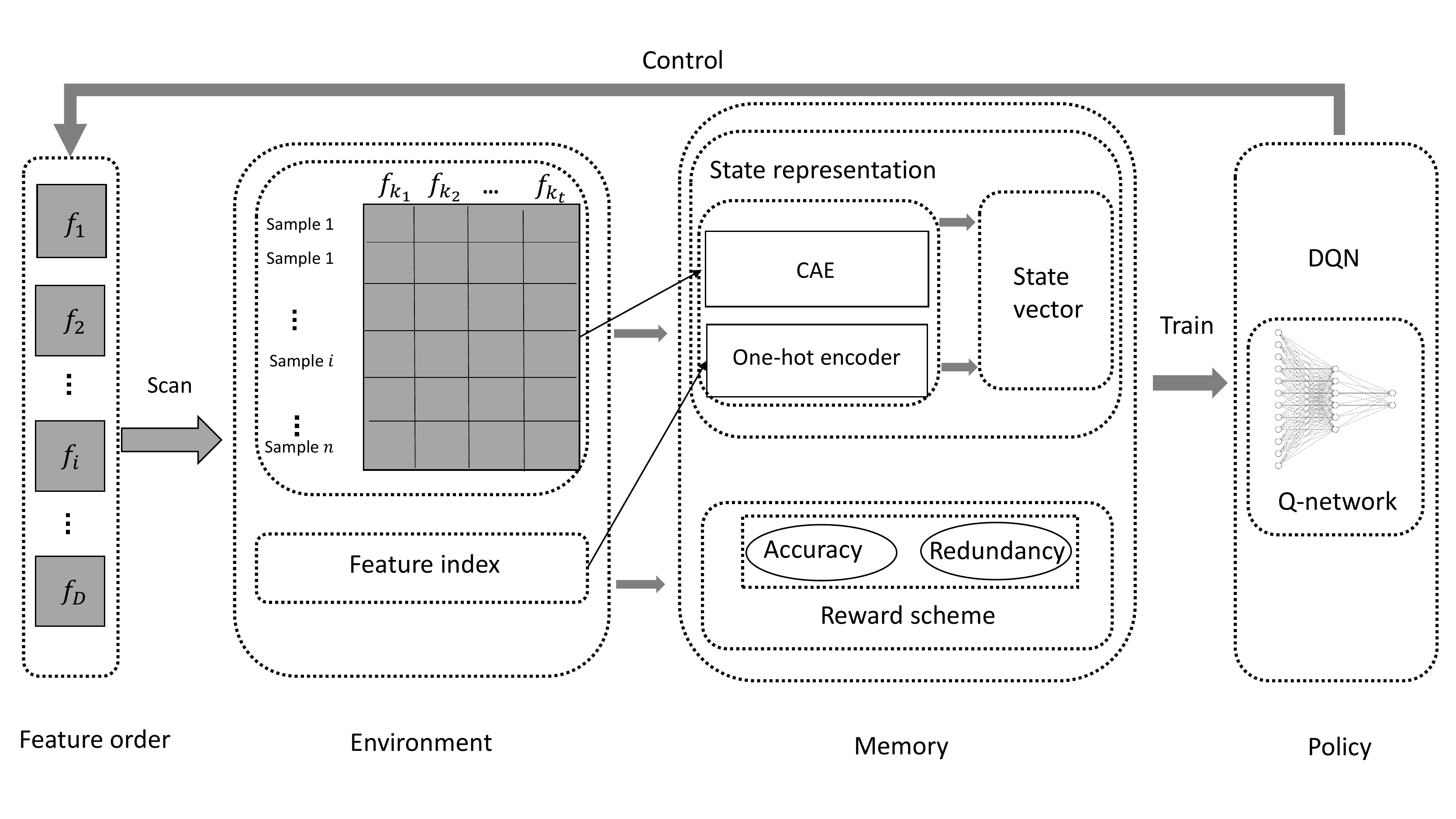}}
\vspace{-0.6cm}
\caption{Framework. Agent scans every feature in feature order. According to the state representation of current environment, agent controls the selection or deselection of the feature. DQN train the policy network.}
\label{fig1}
\end{figure*}

\section{Problem formulation}
We begin this section by introducing some basic notations. A data set of $n$ samples $X = (x_{1}, x_{2}, ..., x_{n})$. Here, a sample $x_{i}$ is a vector with $D$ dimensions (features) $x_{i}= (f_{1}, f_{2}, ..., f_{D})^T$. Let $f_{1}, f_{2}, ..., f_{D}$ denote the $D$ features of $X$. The class label $y = (y_{1}, y_{2}, ..., y_{n})$ are given. Feature selection is defined as selecting a subset of $m$ features from a set of $D$ features with the condition $m \leq D$, that maximizes the predictive accuracy of the learning algorithm.

Automatic feature selection can be regard as a scanning process, where we can sequentially scan the features and decide whether the feature should be selected or deselected. However, the scanning scheme can lead to local optima because of the "nesting effect", where a feature that is selected or discarded can't be selected or discarded in later stages. i.e., sequential forward selection \cite{b10}. Reinforcement learning can maximize the long-term cumulative reward by interacting with the dynamic selected feature subset and keeping exploitation and exploration trade-off.
The nature of reinforcement learning can help to avoid the nesting effect and globally explore the feature space in the scanning scheme. So, in this paper, we design a scanning-based single-agent reinforcement learning framework for feature selection task. The common elements in reinforcement learning are reformulated as follows:
\begin{itemize}
\item\textbf{Environment.}
The environment is a key element in reinforcement learning, which interacts with the agent and provides reward as feedback. In this paper, the environment is responsible for observing the selected features from original features and the current scanning position. In other word, the selected feature subset and the current scanning position are regarded as the environment, where the environment changes dynamically with the action execution of selecting or deselecting feature.
\item\textbf{Agent.}
The agent is also called the scanning agent, which sequentially scans features and makes a decision.
\item\textbf{State space.}
The optimal actions for feature selection is determined by observing the current state of environment. The state ($s$) is to describe the selected feature subset and the current scanning position. Here, we explore a deep representation learning method, convolutional auto-encoder to represent the selected feature subset. 
For the current scanning position, we use index information of the feature, which is scanned by the agent. In this way, the agent identifies which feature it is selecting.
Here, we incorporate the one-hot encoded vector of current feature index into state space.
\item\textbf{Action space.}
The actions of the scanning agent significantly impacts the performance of the overall environment. For feature selection problem, the action $a$ is to make the selection decision for the current feature. So, possible actions are selecting ($a = 1$) or deselecting ($a = 0$) the current feature.

\item\textbf{Reward.}
The reward gives the scanning agent a feedback about how good its action is. It is necessary to provide an immediate reward ($r$) generated by the current selected feature subset. In this paper, we incorporate the predictive accuracy of the selected feature subset and the global redundancy of current feature into the reward function.
\end{itemize}

\section{Proposed Method}

In this section, we first introduce the overview of proposed feature selection framework. Next, we show how to decide the scanning order, how to design reward scheme, and how to employ proposed convolutional auto-encoder for the state representation. Finally, we introduce the scanning-based single-agent reinforcement learning feature selection algorithm in details.

\subsection{Framework Overview}
Fig.~\ref{fig1} presents the overview of our proposed feature selection framework, where a scanning based single-agent reinforcement learning automatically explores feature space to find the optimal feature subset.

This framework mainly includes several steps. At first, we use a relevance-based scanning strategy to decide the scanning order. Then, agent scans features one by one in this scanning order and decides to select or delete this feature according to current state. Here, we use selected feature subset and index information of scanning feature to describe environment. In addition, a proposed convolutional auto-encoder and the one-hot encoding method are utilized to represent the environment. After executing this action, we employ a novel reward scheme which integrates predictive accuracy of selected feature subset with feature redundancy to calculate the reward of this action. Finally, we use Deep Q-learning Network (DQN) \cite{b21} to train the policy for feature selection.

\begin{figure*}[t]
\vspace{0.1cm}
\centerline{\includegraphics[width=18cm]{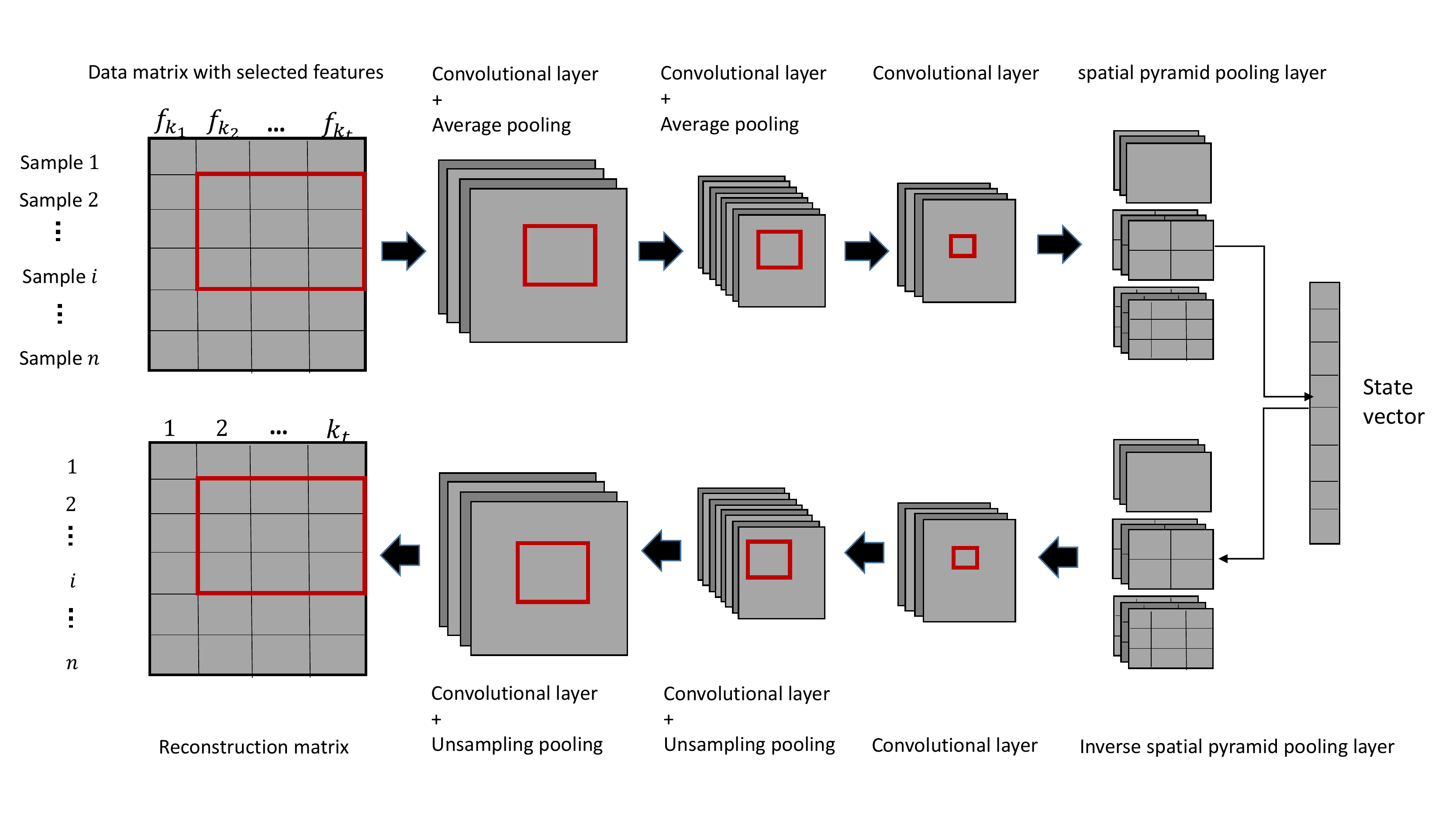}}
\vspace{-0.7cm}
\caption{The architecture of proposed convolutional auto-encoder.}
\label{fig2}
\end{figure*}

\subsection{Relevance-Based Scanning Strategy}
In classification problem, the relevancy of features to target class has a great impact on predictive performance. Considering this, we develop a relevance-based scanning strategy, where we employ feature relevance to decide the scanning order. In other words, we use the ranking based feature relevance to primarily explore relevant feature. Specifically, we calculate the relevance between features and the target class. Then, according to the scores of feature relevance, we sort features in descending order. Finally, the feature ranking is used as the scanning order of the agent. {Here, feature relevance are quantified by information gain \cite{b7}, which measures the amount of information obtained for category prediction by knowing the presence or absence of a feature.} Information gain is calculated based on the information-theoretical concept of entropy, a measure of the uncertainty of a random variable. Let $c \in{\{c_{i}\}_{i = 1}^{L}}$ denote the class variable taking on values from the set of the class categories. The entropy of the class is defined as:
\begin{equation}
H(c) = -\sum_{i}{p(c_{i}) \log_{2}{(p(c_{i}))}},
\end{equation}
where $p(c_{i})$ is the prior probability of the class value $c_{i}$. The conditional entropy of class given the feature $f_{k}$ is calculated as
\begin{equation}
H(c|f_{k})=-\sum_{j}{p(f_{k}^{j})\sum_{i}{p(c_{i}|f_{k}^{j}) \log_{2}{(p(c_{i}|f_{k}^{j}))}}},
\end{equation}
where $p(c_{i}|f_{k}^{j})$ is the posterior probability of $c_{i}$ given the value $f_{k}^{j}$ of feature $f_{k}$. Information gain (IG) can be given by:
\begin{equation}
    IG(f_{k}) = H(c) - H(c|f_{k}).
\end{equation}

\subsection{Reward Scheme}
To capture the interaction between features and the performance of selected feature subset in automatic feature selection process, our reward scheme integrates global redundancy of the feature with the predictive accuracy.

\textbf{Predictive accuracy.} For the selected feature subset $\mathbb{E}$, the predictive accuracy $AC_{\mathbb{E}}$ are calculated based on a predefined classifier. {The predictive accuracy can guide the scanning agent to explore high-quality feature subsets by giving it a positive feedback.}

\textbf{Feature redundancy.} The redundancy of a feature is quantified by global redundancy of this feature, which is the average value of the redundancy of this feature and other features. The feature redundancy can guide the agent to select the lowly redundant features by giving the agent a negative feedback. Here, we use the absolute value of Pearson correlation coefficient to quantify the redundancy between features. Pearson correlation coefficient \cite{b6} is a correlation statistic to calculate the linear correlation between two features. Formally, the redundancy $Rd_{f_{k}}$ of a feature $f_{k}$ is defined as:
\begin{equation}
    Rd_{f_{k}} = \frac{\sum_{j}{|\rho(f_{k}, f_{j})|}}{n}
\end{equation}
where $\rho(f_{k}, f_{j})$ denotes the pearson correlation coefficient \cite{b6} between two features $f_{k}$ and $f_{j}$, which is given by:
\begin{equation}
    \rho(f_{k}, f_{j}) = \frac{cov(f_{k}, f_{j})}{\sigma_{f_{k}} \sigma_{f_{j}}}
\end{equation}
where $cov(f_{k}, f_{j})$ denotes the covariance between two features, and $\sigma_{f_{k}}$ and $\sigma_{f_{j}}$ are standard deviations of the feature.

\subsection{Convolutional auto-encoder}
Auto-encoder has been widely used for representation learning in an unsupervised manner \cite{b24}, where it takes a high dimensional data as input, and outputs a low-dimensional representation vector by minimizing the reconstruction loss between the input and the output. An auto-encoder consists of an encoder which maps the input data to the latent representation vector and a decoder which reconstructs the input data with the latent representation vector.

Considering that the policy network of reinforcement learning requires a fixed-length state representation vector, we need to further extend auto-encoder to represent the selected feature subset, which is dynamically changing in feature space exploring process. Recently, convolution neural network (CNN) \cite{b25} with spatial pyramid pooling layers has shown powerful ability in feature extraction for the multi-size images \cite{b23}. Specifically, the spatial pyramid pooling layer divides every feature map obtained by the last convolutional layer of CNN into multi-level spatial bins with size proportional to the image size. So, the number of bins is fixed regardless of the feature map size. Then, this layer concatenates these bins to generate the fixed-length vector. In this paper, we combine the CNN and auto-encoder named convolutional auto-encoder (CAE) to obtain accurate environment state representation of fixed length.

The architecture of proposed convolutional auto-encoder is shown in Fig.~\ref{fig2}. Here, we regard the dataset $X$ (a data matrix of $n$ samples with the selected $m$ features) as an image. In proposed CAE network structure, the last layer of encoder is replaced with the spatial pyramid pooling layer to generate the fixed-length representation. In the meantime, we design an inverse spatial pyramid pooling layer as the first layer of decoder. The detail operation of the symmetric convolutional auto-encoder is shown as follows:
\paragraph{Encoder}the input image $X$ is subjected to the operation of the convolution layers to obtain a set of convolution feature maps, and sequentially passed through the average pooling layers. Here, suppose that the last convolutional layer of the encoder obtain $H$ feature maps. They are the input to a spatial pyramid pooling layer, where it pools every feature map in multi-level local spatial bins, i.e., $1\times{1}$, $2\times{2}$, and $3\times{3}$ bins of sizes proportional to the feature map size. Then, the spatial pyramid pooling layer performs average pooling on each bin, and concatenates these bins to form a fixed-dimensional latent representation vector $z_{1}$. Note that all convolutional layers in this paper have zero-padding added to ensure that each convolutional layer outputs the same size as the input.
\paragraph{Decoder}based on the latent representation vector $z_{1}$, the decoder reconstructs as closely as possible the original input data. The decoder architecture is a reverse mirror of the encoder. In this paper, we design an inverse spatial pyramid pooling layer. Specifically, the fixed-length latent representation vector $z_{1}$ will be reshaped into $H$ feature maps of size $1\times{1}$, $H$ feature maps of size $2\times{2}$, and $H$ feature maps of size $3\times{3}$. Then, upsample feature maps of every size back to the same size as the input of spatial pyramid pooling layer. The average value of the $i$-th expansion feature maps from three sizes is regarded as a feature map. Finally, the $H$ feature maps from the inverse spatial pyramid pooling layer are inputted into the upsampling layers followed by the convolutional layers. The filters of Convolutional layers of the decoder correspond the filters of Convolutional layers of the encoder. Here, upsampling layers are used to upsample the feature maps back to original size. Therefore, the size of output of the decoder is the same as that of input image of encoder.

Finally, the fixed-length latent representation vector $z_{1}$ learned by proposed convolutional auto-encoder is used to be  part of the state representation in reinforcement learning.

\begin{algorithm}[htbp]
    \small
    \caption{Scanning Based Single-Agent Reinforcement Learning Feature Selection}
    \label{alg:DRLFS}
    \begin{algorithmic}[1]
        \STATE Initialize replay memory $M$ with capacity $P$; 
        \STATE Initialize policy network $Q$ and target network $\hat{Q}$ with random parameters $\theta$ and $\hat{\theta}$ respectively with respect to the normal distribution;
        \STATE initialize the selected feature subset $\mathbb{E}$ and the scanning pointer $i$;
        \STATE Construct the scanning order $F$ by the relevance-based scanning strategy;
        \WHILE{$episode \leq K$}
            \FOR{$i = 1$ to $D$}
                \STATE Scanning the feature $f_{i}$ one by one in $F$;
                \STATE Choose a selection or deselection action $a_{t}$ for current state $s_{t}$ based on $\epsilon$-greedy policy strategy;
                \IF{$a_{t}==1$}
                   \STATE Update the feature subset: $ \mathbb{E} = \mathbb{E} \cup f_{i}$;
                \ENDIF
                \STATE Calculate the reward: $r_{t}=AC_{\mathbb{E}}-Rd_{f_{i}}$
                \STATE Learning the latent representation vector $z_{1}$ for data matrix $X$ with new selected feature subset;
                \STATE Use one-hot encoded vector $z_{2}$ to represent the scanning pointer $i$;
                \STATE Update the next state vector: $s_{t+1}=(z_{1},z_{2})$
                \STATE Store transition $(s_{t}, a_{t}, r_{t}, s_{t+1})$ in replay memory $M$;
                \STATE Randomly sample a mini-batch $\mathfrak{D}_{t}$ of transitions $(s_{t}, a_{t}, r_{t}, s_{t+1})$ from memory $M$;
                \STATE Set $y_{t} = r_{t} +\gamma \max_{a_{t+1}} Q(s_{t+1}, a_{t+1} | \hat{\theta})$;
                \STATE Perform a gradient decent step on $(y_{t} - q(s_{t}, a_{t}))^2$ with respect to network parameters $\theta$;
                \STATE Update the target network parameters $\hat{\theta}$ after every $C$ steps: $\hat{\theta} = \theta$;
            \ENDFOR
        \ENDWHILE
    \end{algorithmic}
\end{algorithm}

\subsection{Scanning Based Single-Agent Reinforcement Learning Feature Selection}
As algorithm \ref{alg:DRLFS} shows, we propose a scanning based single-agent reinforcement learning feature selection method. The detailed process is explained as follows:

At first, from line 1 to 3 in algorithm \ref{alg:DRLFS}, we initialize the parameters of deep Q-network, the selected feature subset and the scanning pointer, which points the current scanning feature.

Next, the relevance-based scanning strategy decides the scanning order. Specifically, the relevance-based scanning strategy employs information gain to calculate the correlation between feature and target class (feature relevance), and then constructs the feature scanning order by ranking feature relevance in descending order. 

Thirdly, the agent begins to scan feature space one by one in this feature scanning order. When scanning a feature, agent needs to make a decision about selecting or deselecting this feature based on its policy network, where the inputs are provided by the state vector, and each separate output represents the $Q$-values for selection or deselection action. With $\epsilon$-greedy policy strategy, the agent chooses a selection or deselection action to execute. Specifically, with the policy $\epsilon\in[0,1]$, the agent chooses the greedy action with the highest $Q$-values. With the probability of $1-\epsilon$, the agent chooses a random action. This way helps the agent to balance the exploration and exploitation in automatic feature selection process. 

Fourthly, after executing a selection or deselection action on current selected feature subset, environment will transform to the next state $s_{t+1}$ and give agent a reward value $r_{t}$. At first, environment state $s_{t+1}$ include two parts, e.g., new selected feature subset and the position information of current scanning feature. For the selected feature subset, we adopt our proposed convolutional auto-encoder to generate a fixed-length latent representation vector $z_{1}$. For current scanning position, we use the one-hot encoded vector $z_{2}$ of the feature index to represent. Then, we concatenate the fix-length latent representation vector and one-hot encoded vector to generate the final state vector $s_{t+1}$. The reward value is measured by the predictive accuracy of selected feature subset and feature redundancy. In the reward scheme, agent can search an optimal subset of lowly redundant features for the classification model.

Finally, after accumulating more and more transition tuple $(s_{t}, a_{t}, r_{t}, s_{t+1})$ in memory $M$. Agent will train the neural networks via experience replay independently. The agent randomly samples mini-batches of experiences from memory $M$. Then, based on the mini-batch samples, the agent trains its policy network to obtain the maximum expected long-term return by the Bellman Equation \cite{b16}:
\begin{equation}
Q(s_{t}, a_{t}|\theta) = r_{t} + \gamma \max Q(s_{t+1}, a_{t+1}|\theta)
\end{equation}
where $\theta$ is the parameter set of $Q$-network, and $\gamma$ is the discount factor.

\section{Experiment}
In this section, we validate our proposed method and analyze the performance on the real world data.

\subsection{Datasets}
The effectiveness of our proposed method is evaluated on four publicly available real datasets.

\textbf{Forest Cartographic} shows some cartographic variables to classify forest categories that range from 1 to 7, which is download from Kaggle \footnote{https://www.kaggle.com/c/forest-cover-type-prediction/data}. It is comprised of 15120 samples with 54 features.

\textbf{Amazon Employee Access} is to predict whether employees have access to resources taken from Kaggle \footnote{https://www.kaggle.com/c/amazon-employee-access-challenge/data}. It consist of 32769 samples characterized by 9 integer-valued features, 2 classes.

\textbf{UCI Nomao} collects location data (name, phone, etc.) from many sources to detect what data refer to the same place. It includes 34465 samples with 118 features, 2 classes. Available on the UCI Machine Learning Repository \footnote{https:www.openml.org/d/1486}.

\textbf{Protein Lysine Glycation} is to predict protein lysine glycation sites by some sequence information, which is collected from the Compendium of Protein Lysine Modifications \footnote{http://cplm.biocuckoo.org/}. In this study, we refer to the configuration in \cite{b12}. Finally, this dataset contain 630 samples with 402 features, 2 classes.

\begin{table}[htbp]
\small\vspace{-0.2cm}
\addtolength{\tabcolsep}{5pt}
\caption{Dataset Description}
\begin{center}
\begin{tabular}{|c|c|c|c|}
\hline
\textbf{Dataset} & \textbf{Samples} & \textbf{Features} & \textbf{Classes} \\
\hline
Cartographic &  15120 & 54 & 7 \\
\hline
AmazonEA & 32769 & 9 & 2 \\
\hline
Nomao& 34465 & 118 & 2 \\
\hline
Glycation & 630 & 402 & 2 \\
\hline
\end{tabular}
\end{center}
\label{tab1}
\vspace{-0.5cm}
\end{table}

We present details about the four datasets in Table \ref{tab1}. The Cartographic, AmazonEA, Nomao and Glycation in Table \ref{tab1} represent Forest cartographic, Amazon Employee Access, UCI Nomao and Protein Lysine Glycation dataset, respectively.

\subsection{Evaluation Metrics}
For the comparison of the predictive performance on the benchmark datasets, four widely used indicators were used, i.e., Accuracy (ACC), precision, recall and $F$-measure. Precision is the ratio of true positives to true positives plus false positives. Recall is the ratio of true positives to true positives plus false negatives. $F$-measure is the harmonic mean of precision and recall \cite{b19}.
\begin{equation}
ACC = \frac{TP+TN}{TP+TN+FP+FN}
\end{equation}
\begin{equation}
Precision = \frac{TP}{TP+FP}
\end{equation}
\begin{equation}
Recall = \frac{TP}{TP+FN}
\end{equation}
\begin{equation}
F-measure = \frac{2*P*R}{P+R}
\end{equation}
where $TP$, $TN$, $FP$, $FN$ represent the number of true positives, true negatives, false positives and false negatives respectively.

\begin{table*}[htbp]
\addtolength{\tabcolsep}{4pt}
\caption{Computation Cost}
\vspace{-5mm}
\begin{center}
\begin{tabular}{|c|c|c|c|c|c|c|c|c|c|c|c|c|}
\hline
 & \multicolumn{2}{|c|}{\textbf{Cartographic}} &
\multicolumn{2}{|c|}{\textbf{AmazonEA}} &
\multicolumn{2}{|c|}{\textbf{Nomao}} &
\multicolumn{2}{|c|}{\textbf{Glycation}} \\
\cline{2-3} 
\cline{4-5} 
\cline{6-7} 
\cline{8-9} 
 & \textbf{\textit{CPU}}  & \textbf{\textit{Memory(MB)}}
&\textbf{\textit{CPU}}  & \textbf{\textit{Memory(MB)}}
&\textbf{\textit{CPU}} & \textbf{\textit{Memory(MB)}}
&\textbf{\textit{CPU}} & \textbf{\textit{Memory(MB)}}\\
\hline
MARLFS& 74$\%$ & 1525 & 97$\%$ & 1485 & 82$\%$ & 1649 & 70$\%$ & 1579\\
\hline
SADRLFS & 56$\%$ & 1516 & 42$\%$ & 1458 & 48$\%$ & 1612 & 30$\%$ & 1478\\
\hline
\end{tabular}
\label{tab3}
\end{center}
\vspace{-4mm}
\end{table*}

\begin{table*}[htbp]
\addtolength{\tabcolsep}{1.3pt}
\caption{Overall accuracy on different classifiers}
\vspace{-5mm}
\begin{center}
\begin{tabular}{|c|c|c|c|c|c|c|c|c|c|c|c|c|}
\hline
 & \multicolumn{3}{|c|}{\textbf{Cartographic}} &
\multicolumn{3}{|c|}{\textbf{AmazonEA}} &
\multicolumn{3}{|c|}{\textbf{Nomao}} &
\multicolumn{3}{|c|}{\textbf{Glycation}} \\
\cline{2-4} 
\cline{5-7} 
\cline{8-10} 
\cline{11-13} 
 & \textbf{\textit{RF}} & \textbf{\textit{XGB}} & \textbf{\textit{DT}}
&\textbf{\textit{RF}} & \textbf{\textit{XGB}} & \textbf{\textit{DT}}
&\textbf{\textit{RF}} & \textbf{\textit{XGB}} & \textbf{\textit{DT}}
&\textbf{\textit{RF}} & \textbf{\textit{XGB}} & \textbf{\textit{DT}}\\
\hline
ReliefF& 0.6164 & 0.6052 & 0.6402 & 0.9005 & 0.9426 & 0.9277 & 0.8468 & 0.7508 & 0.8468 & 0.6825 & 0.7302 & 0.6824\\
\hline
mRMR & 0.5291 & 0.5569 & 0.5423 & 0.9027 & 0.9426 & 0.9282 & 0.8689 & 0.8587 & 0.8724 & 0.55556 & 0.6825 & 0.5873\\
\hline
Chi-square & 0.7090 & 0.6131 & 0.7586 & 0.8953 & 0.9426 & 0.9270 & 0.8886 & 0.8587 & 0.8840 & 0.6825 & 0.6825 & 0.6508\\
\hline
SFS & 0.7368 & 0.6190 & 0.7579 & 0.9200 & \textbf{0.9429} & 0.9286 & 0.9101 & 0.8616 & 0.9156 & 0.6508 & 0.6984 & 0.7460\\
\hline
BDEFS & 0.7169 & 0.6409 & \textbf{0.7685} & 0.9335 & 0.9426 & 0.9362 & 0.9449 & 0.9269 & 0.9442 & 0.6825 & 0.6190 & 0.5556\\
\hline
SARLFS & 0.7989 & 0.6356 & 0.7619 & 0.9381 & 0.9411 & 0.9212 & 0.9605 & 0.9298 & 0.9475 & 0.6825 & 0.6508 & 0.6667 \\
\hline
MARLFS & 0.8102 & 0.6138 & 0.7679 & 0.9463 & 0.9423 & 0.9219 & 0.9620 & 0.9283 & 0.9510 & 0.7619 & 0.6667 & 0.6349\\
\hline
SADRLFS & \textbf{0.8697} & \textbf{0.7725} & 0.7606 & \textbf{0.9512} & 0.9426 & \textbf{0.9478} & \textbf{0.9707} & \textbf{0.9579} & \textbf{0.9527} & \textbf{0.7778} & \textbf{0.7619} & \textbf{0.7460}\\
\hline
\end{tabular}
\label{tab2}
\end{center}
\end{table*}

\subsection{Baselines}
We compare our proposed method with seven baselines, including both traditional feature selection methods and reinforcement learning framework for feature selection
\begin{itemize}
\item\textbf{ReliefF.} The idea of ReliefF is to estimate the quality of features based on their values to distinguish between the samples that are close with each other \cite{b26}. In this study, we select the $K$ highest ReliefF scores, where $K$ is equal to the number of selected features in our proposed method.
\item\textbf{Maximum Relevance Minimum Redundancy (mRMR).} The mRMR method selects features that have the highest relevance with the class label and are also maximally dissimilar to each other \cite{b8}. In this experiments, we select the $K$ highest mRMR scores.
\item\textbf{Chi-square Scores.} Chi-square scores \cite{b30} try to measure the degree of independence between the feature and target class. In this experiments, we select the $K$ highest chi-square scores.
\item\textbf{Sequential Forward Selection (SFS).} Sequential forward selection (SFS) algorithm \cite{b10} is a bottom-up search procedure which begin from an empty set and iteratively add the next best features to the set until a predefined number of features are selected. In this experiment, we select the feature subset of the highest accuracy.
\item\textbf{Binary Differential Evolution Algorithm for Feature Selection (BDEFS).} Binary differential evolution algorithm search the optimal feature subset by a binary mutation operator and a binary crossover operator. Here, we refer to the algorithm framework and parameter settings of the paper \cite{b12}.
\item\textbf{Single-Agent Reinforcement Learning for Feature Selection (SARLFS).}
SARLFS \cite{b18} is a model-based reinforcement learning, where the agent learns a model of the environment, represented as a dynamic Bayesian network that describes rewards and state transitions, and efficiently computes a policy using dynamic programming.
\item\textbf{Multi-Agent Reinforcement Learning for Feature Selection (MARLFS).}
MARLFS \cite{b19} automatically explores feature subsets by regarding each feature as an agent. A feature agent decide to select or deselect a feature. The reward scheme guides the cooperation and competition between agents by integrating feature-feature mutual information with accuracy.
\end{itemize}

\subsection{Overall Performances}
In this experiment, we compare our proposed method (SADRLAFS) with different baseline methods based on random forest (RF) \cite{b27} with 100 decision trees. For proposed convolutional auto-encoder network, we use epochs of 10, and a learning rate of 0.005. The inputs of convolutional auto-encoder are scaled $[-1, 1]$. The convolutional layers are set to 16, 32, 16, 32, 16, and 1 filters with $3\times{3}$ convolution kernel, sequentially. Tanh function is used as the activation function in the convolution layers. The pooling layers perform average pooling with $2\times{2}$ kernel. The spatial pyramid pooling layer performs 4-level pyramid: $\{1\times{1}$, $2\times{2}$, $3\times{3}$, $4\times{4}\}$ (totally 30 bins). The reinforcement learning algorithm performs with discount factor $\gamma$ of 0.9, a learning rate of 0.01, and mini-batch size of 32. For experience replay, we use memory size of 400. Then, we set the Q network as a one-layer ReLU with 100 nodes in the hidden layer. The Adam Optimizer is adopted to optimize the parameters of all neural network structure. In this paper, $90\%$ of sample instances are used for training sets and $10\%$ of sample instances are used for validation sets.

\begin{figure*}[htbp]
\centerline{
\subfigure[Accuracy]{
\includegraphics[width=4cm]{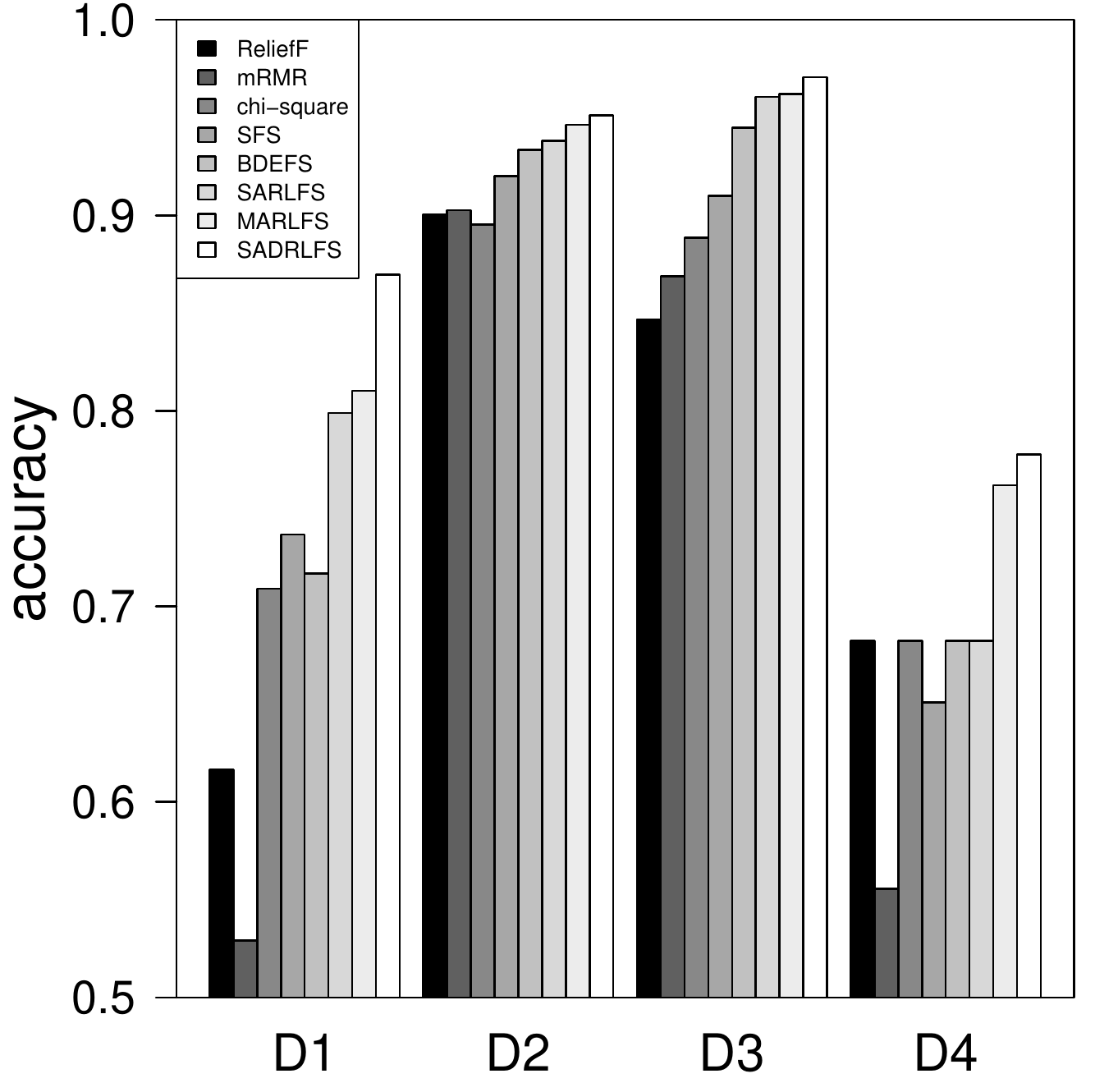}
}
\hspace{0in}
\subfigure[Precision]{
\includegraphics[width=4cm]{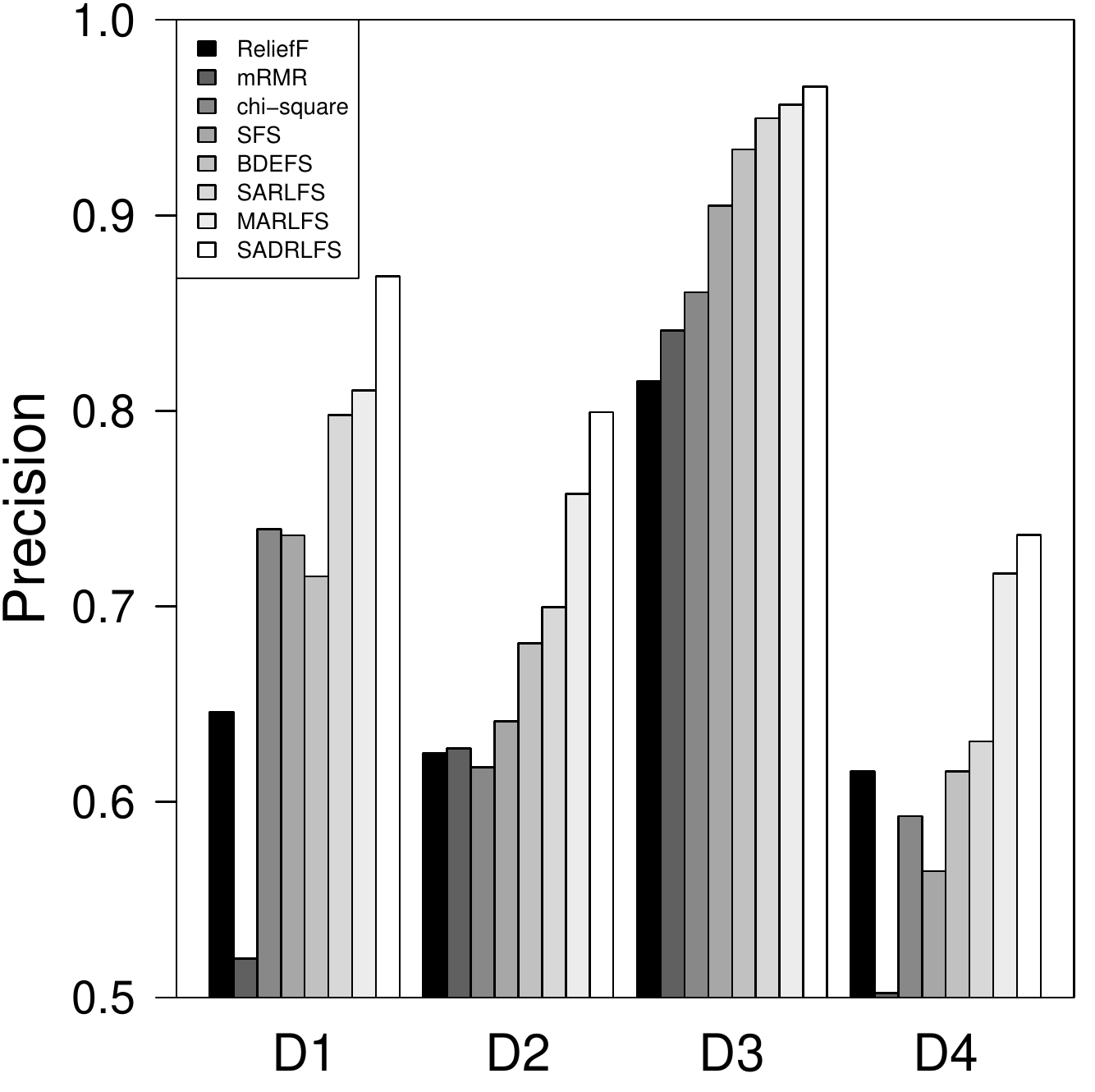}
}
\hspace{0in}
\subfigure[Recall]{
\includegraphics[width=4cm]{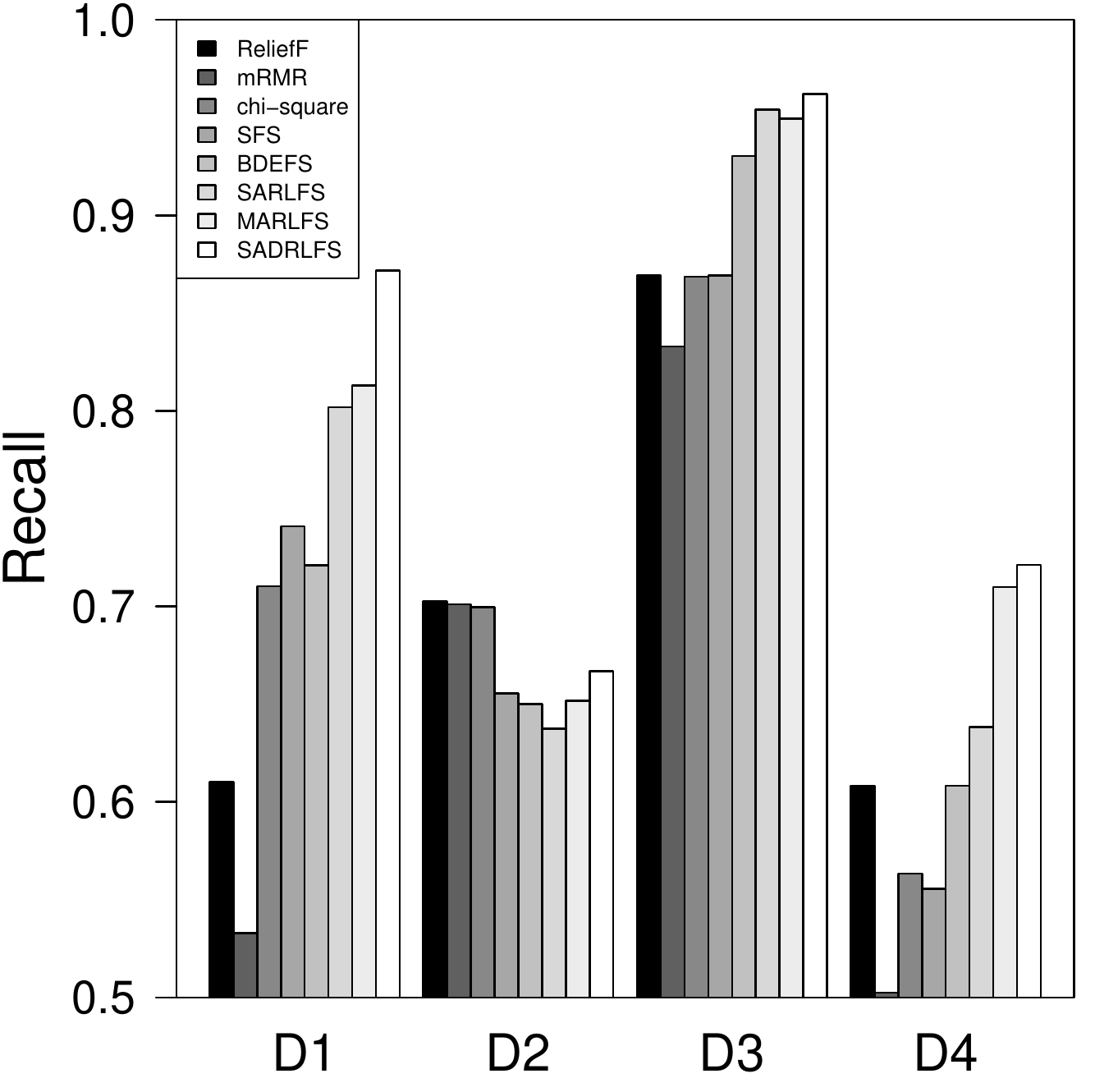}
}
\hspace{0in}
\subfigure[F-Measure]{
\includegraphics[width=4cm]{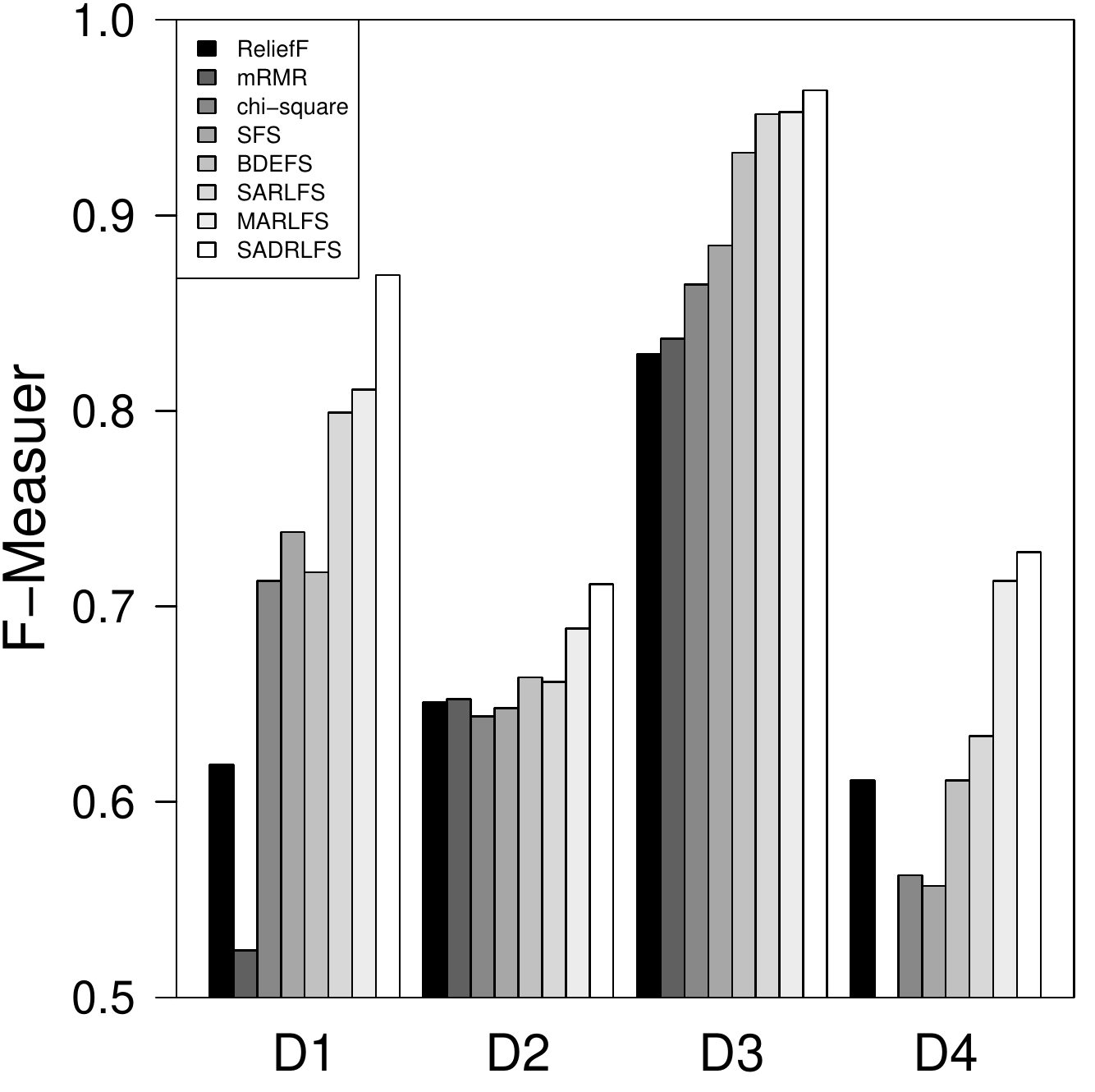}
}
}
\vspace{-2.5mm}
\caption{Performance comparison of different feature selection algorithms.}
\label{fig3}
\vspace{-3mm}
\end{figure*}

At first, we compare our proposed single-agent reinforcement learning framework (SADRLFS) with multi-agent reinforcement learning (MARLFS). From Table \ref{tab3}, we can see that proposed single-agent reinforcement learning takes less computation cost than multi-agent reinforcement learning for feature selection. This is because for multi-agent reinforcement learning, the exponential growth of state-action space in the number of state and action variables leads directly to the high computational complexity. In addition, comparison with seven baselines, the classification performance on four real-world datasets are shown in Fig.~\ref{fig3}, where D1, D2, D3 and D4 represent the Cartographic, AmazonEA, Nomao and Glycation datasets. Our proposed method obtains better performance on four datasets.

Aside from feature selection, it is known that Classifiers play an import role in classification accuracy. So, in this experiment, we also adopt another two most popular classifiers, i.e., XGBoost \cite{b28} and Decision Tree (DT) \cite{29} to investigate the robustness of our selected feature subset. Table \ref{tab2} shows the results of comparison on different classifiers. The bold values indicate the best performance in the table. For Gartographic dataset, the accuracy (0.7606) of our algorithm is slightly lower than BDEFS (0.7685) on DT classifier. For AmazonEA dataset, the accuracy (0.9426) of our method also slightly lower than SFS (0.9429) on XGBoost classifier. In general, our method obtains better performance on most datasets in terms of overall accuracy.

\begin{figure*}[htbp]
\centerline{
\subfigure[Accuracy]{
\includegraphics[width=4cm]{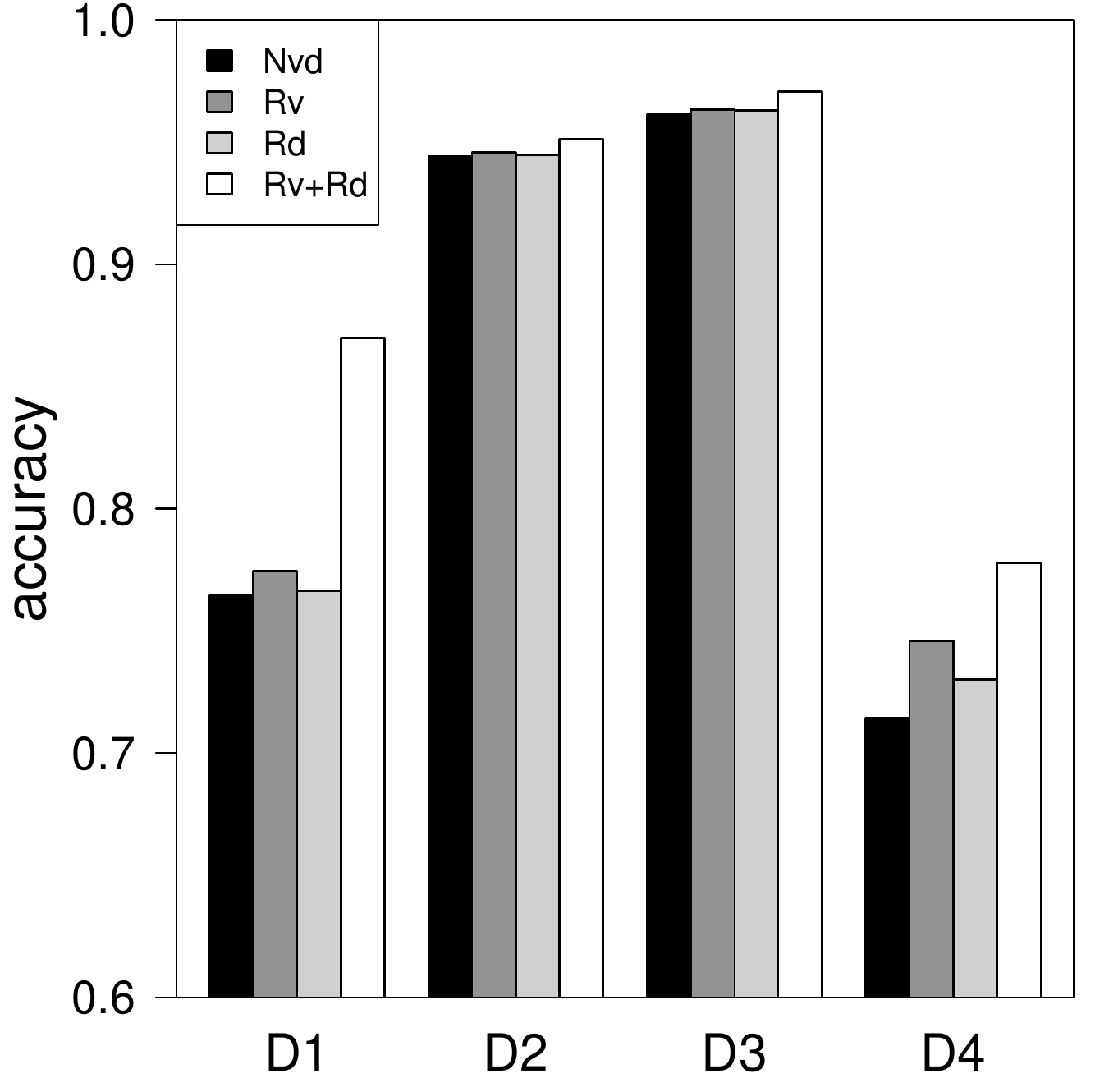}
}
\hspace{0in}
\subfigure[Precision]{
\includegraphics[width=4cm]{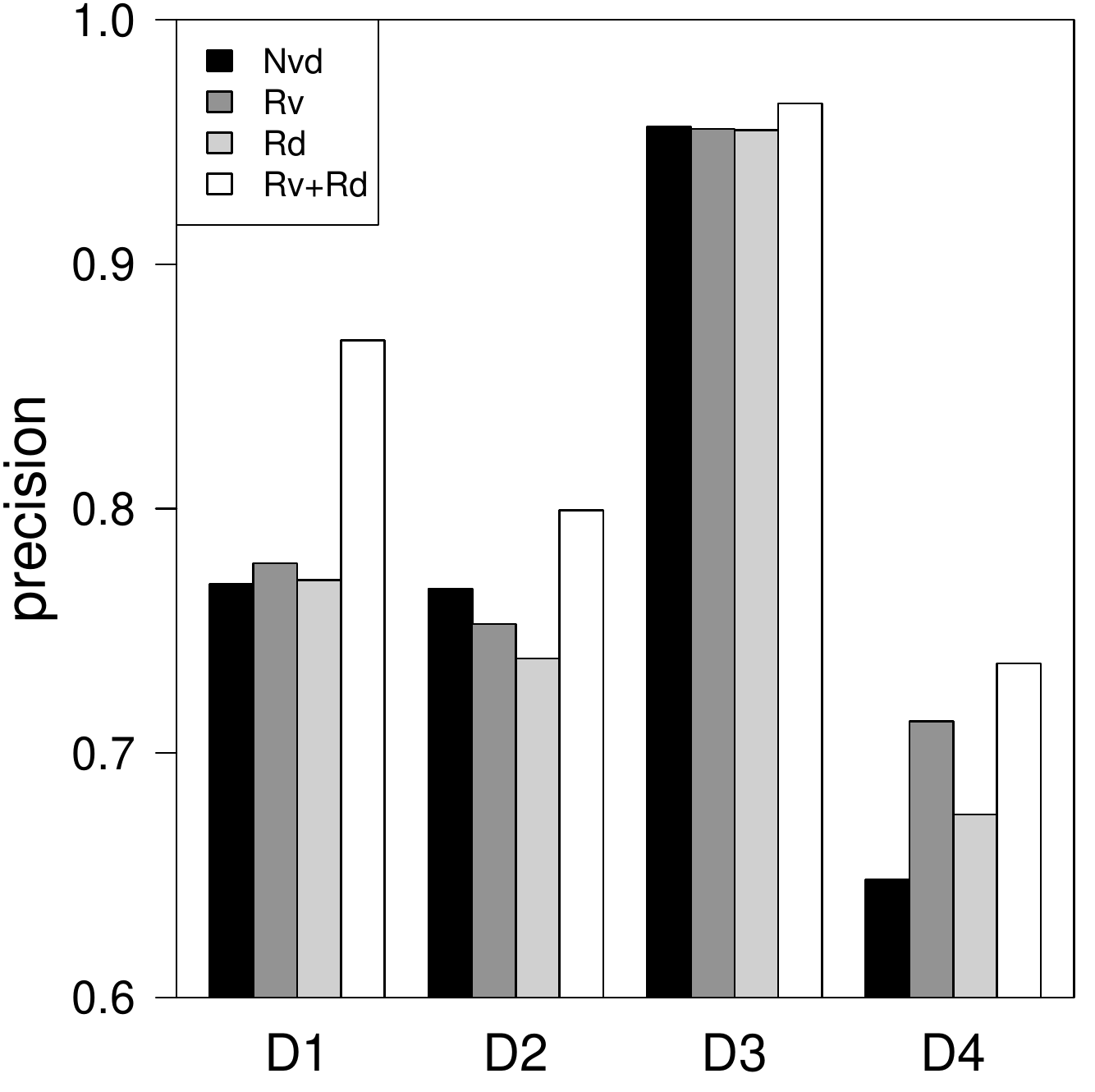}
}
\hspace{0in}
\subfigure[Recall]{
\includegraphics[width=4cm]{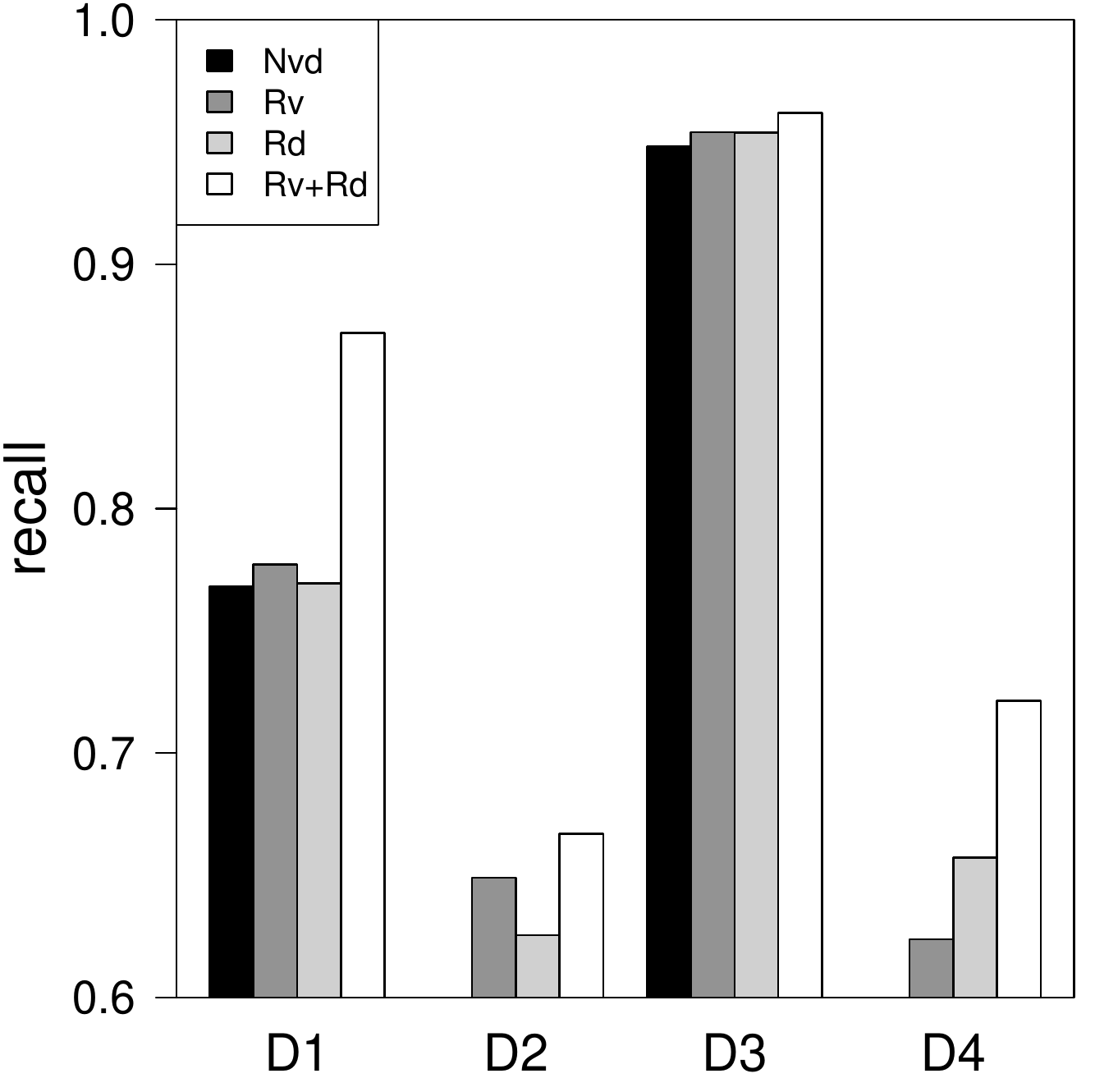}
}
\hspace{0in}
\subfigure[F-Measure]{
\includegraphics[width=4cm]{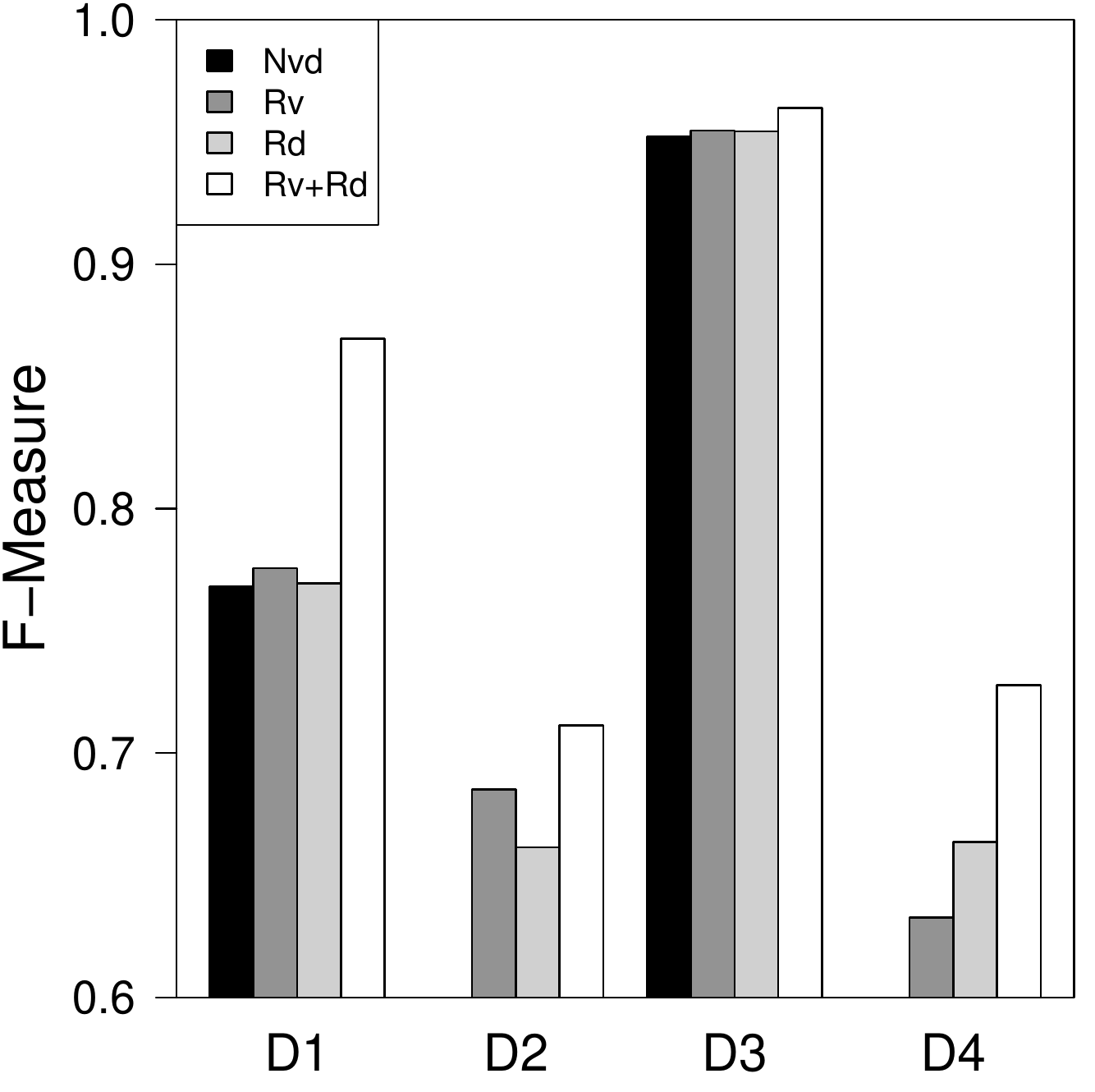}
}
}
\vspace{-2.5mm}
\caption{Study of incorporating relevance and redundancy.}
\label{fig4}
\end{figure*}

\subsection{Study of incorporating relevance and redundancy}
In this experiment, we further study the impact of incorporating feature relevance and redundancy. Here, incorporating relevance is realized by the scanning order, which is the ranking of information gain scores between features and class. Incorporating redundancy is realized by integrating pearson correlation coefficient between features into the reward function. We investigate four cases: (i) Nvd that doesn't incorporate feature relevance in the scanning order, and feature redundancy in reward function; (ii) Rv that only incorporates feature relevance in the scanning order; (iii) Rd that only incorporates feature redundancy in reward function; (iv) Rv+Rd that incorporates feature relevance and redundancy. The experiment performance on four cases are shown in Fig.~\ref{fig4}. We can see that the combination between feature relevance with feature redundancy can improve the predictive performance. This might be explained by complex multi-way interaction among features, where a lowly relevant feature could significantly improve the predictive performance if it is used together with other complementary features, and a highly relevant feature may become redundant when used together with some features.

\subsection{Study of Environment Representation}
In order to widely study the performance of our proposed algorithm, this part shows the comparative results with environment representation.

\begin{figure*}[htbp]
\centerline{
\subfigure[Accuracy]{
\includegraphics[width=4cm]{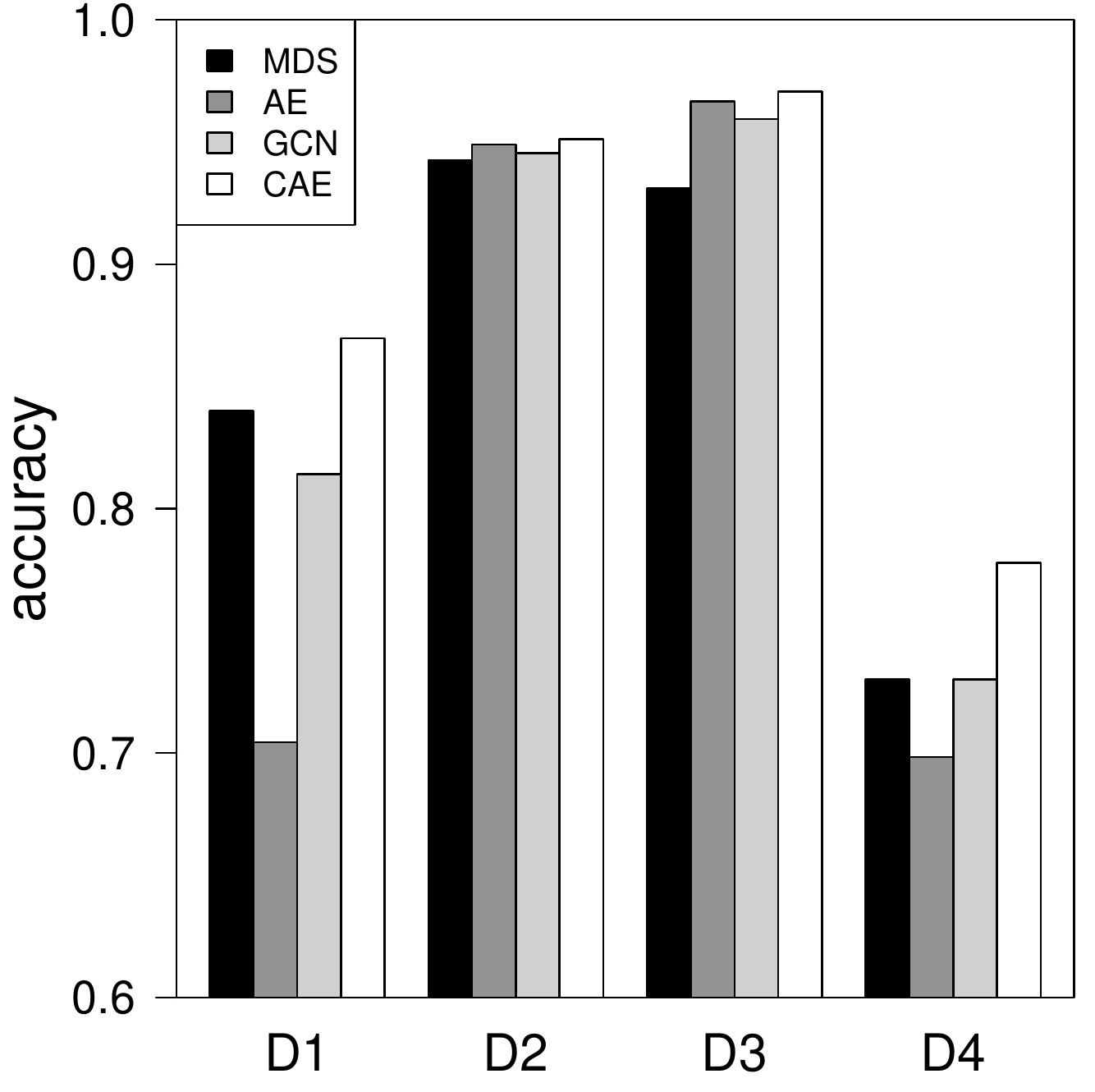}
}
\hspace{0in}
\subfigure[Precision]{
\includegraphics[width=4cm]{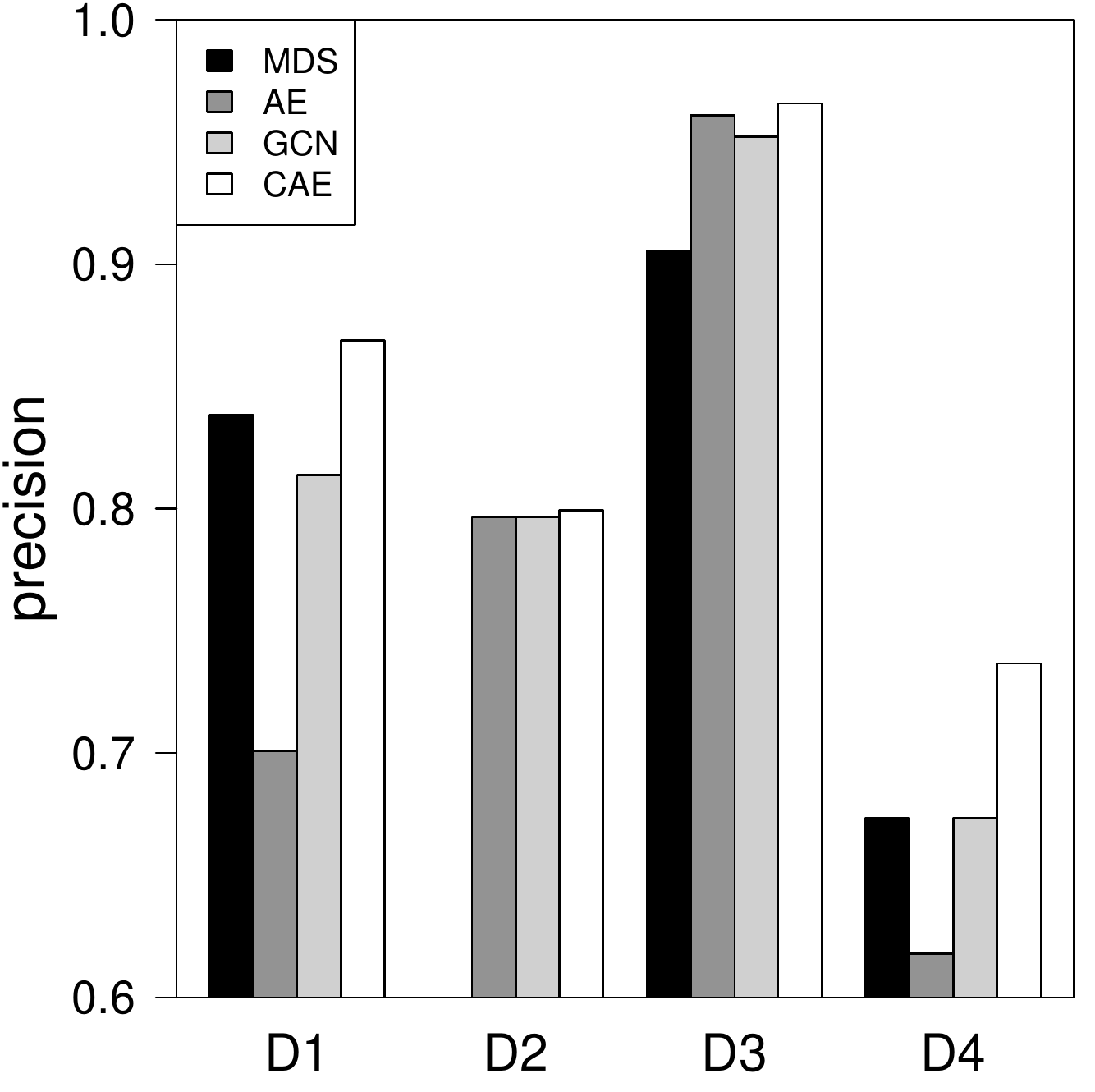}
}
\hspace{0in}
\subfigure[Recall]{
\includegraphics[width=4cm]{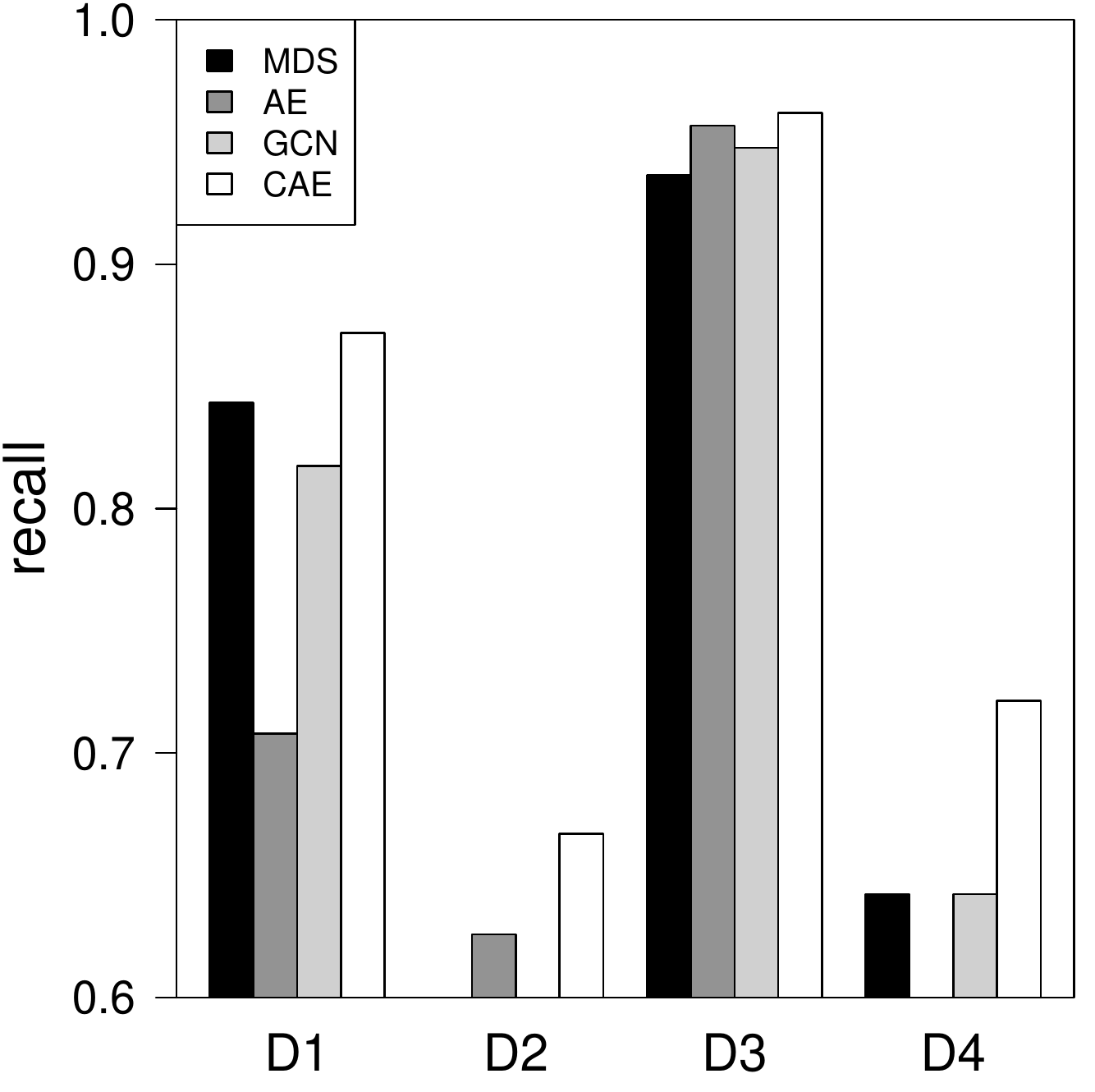}
}
\hspace{0in}
\subfigure[F-Measure]{
\includegraphics[width=4cm]{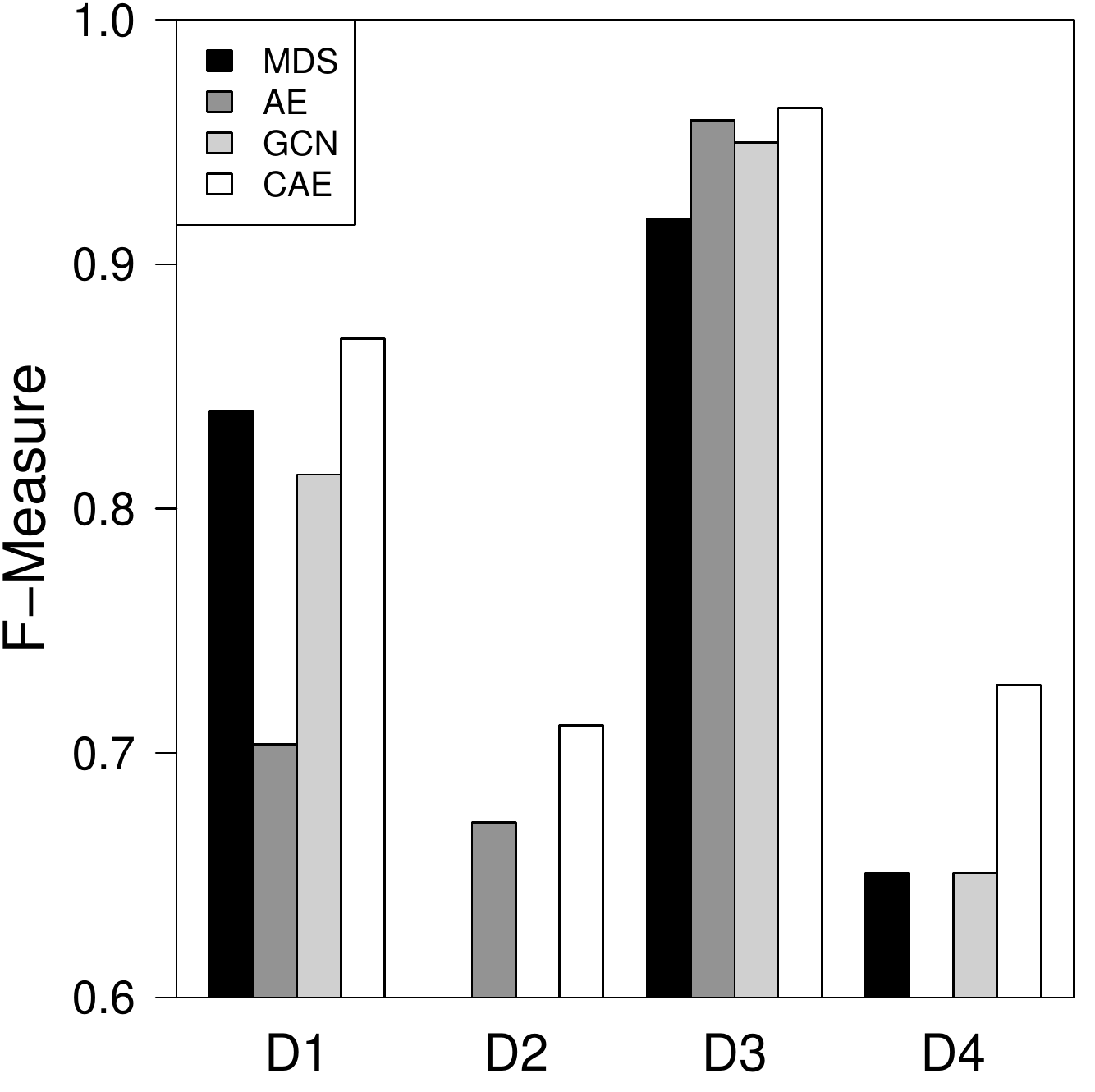}
}
}
\vspace{-2.5mm}
\caption{Study of first state representation.}
\label{fig5}
\vspace{-3mm}
\end{figure*}

At first, we investigate the first part of environment state, which is the representation of data matrix with selected feature subsets. We investigate four representation learning method, i.e., (i) MDS: meta descriptive statistics including the standard deviation, minimum value, maximum value, the first quartile, the second quartile, and the third quartile; (ii) AE: auto-encoder based representation learning method, which applies two-step auto-encoder learning to representation dynamic data matrix; (iii) GCN: Graph convolutional network (GCN) based representation learning, which adopts GCN to embedding graph node of dynamic feature correlation graph; (iv) CAE: Proposed convolutional auto-encoder based representation learning. The representation learning methods of MDS, AE, GCN refer to \cite{b19}. The results are shown in Fig.~\ref{fig5}. From this figure, CAE obtains the best performance.

\begin{figure*}[htbp]
\centerline{
\subfigure[Accuracy]{
\includegraphics[width=4cm]{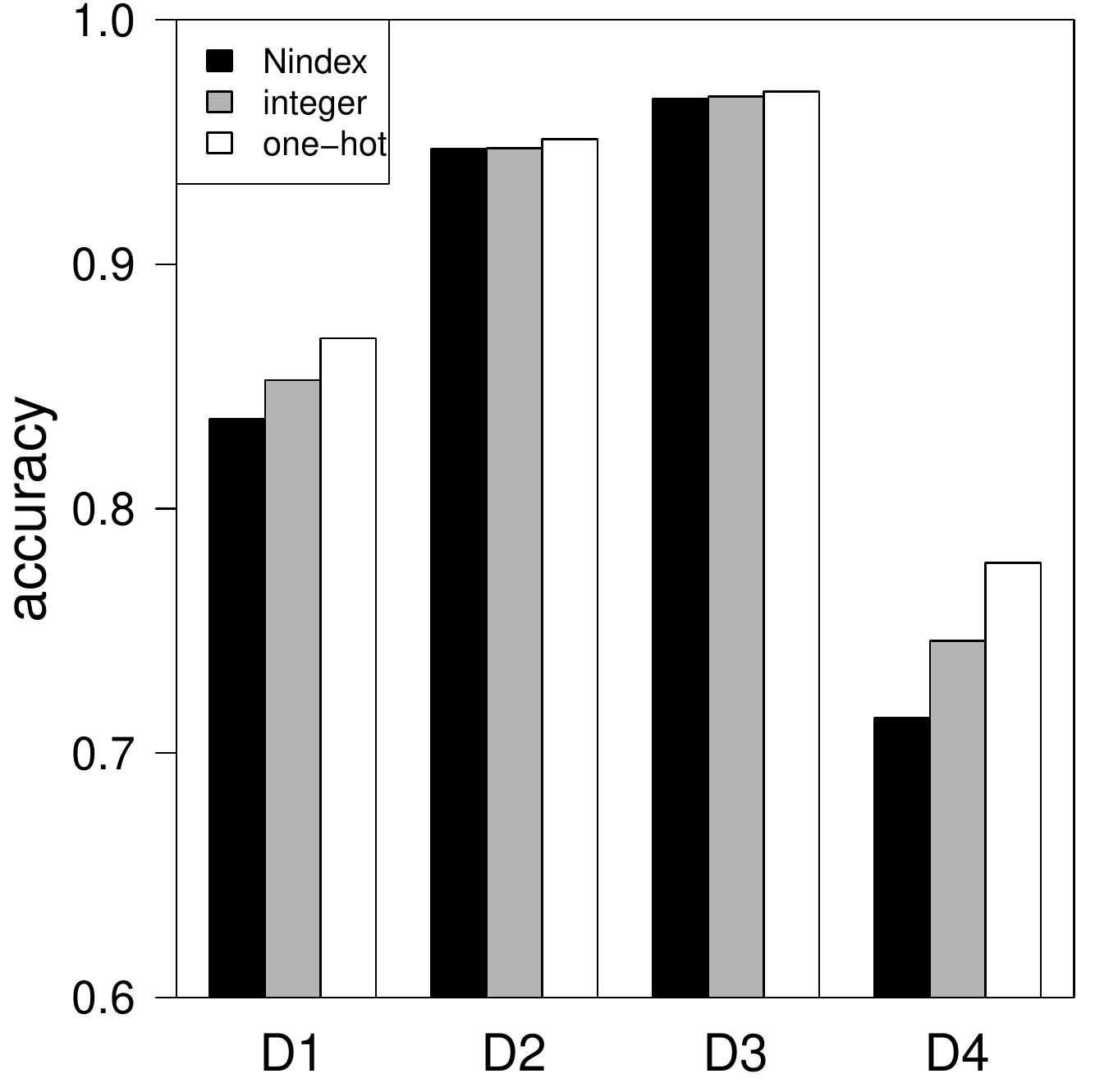}
}
\hspace{0in}
\subfigure[Precision]{
\includegraphics[width=4cm]{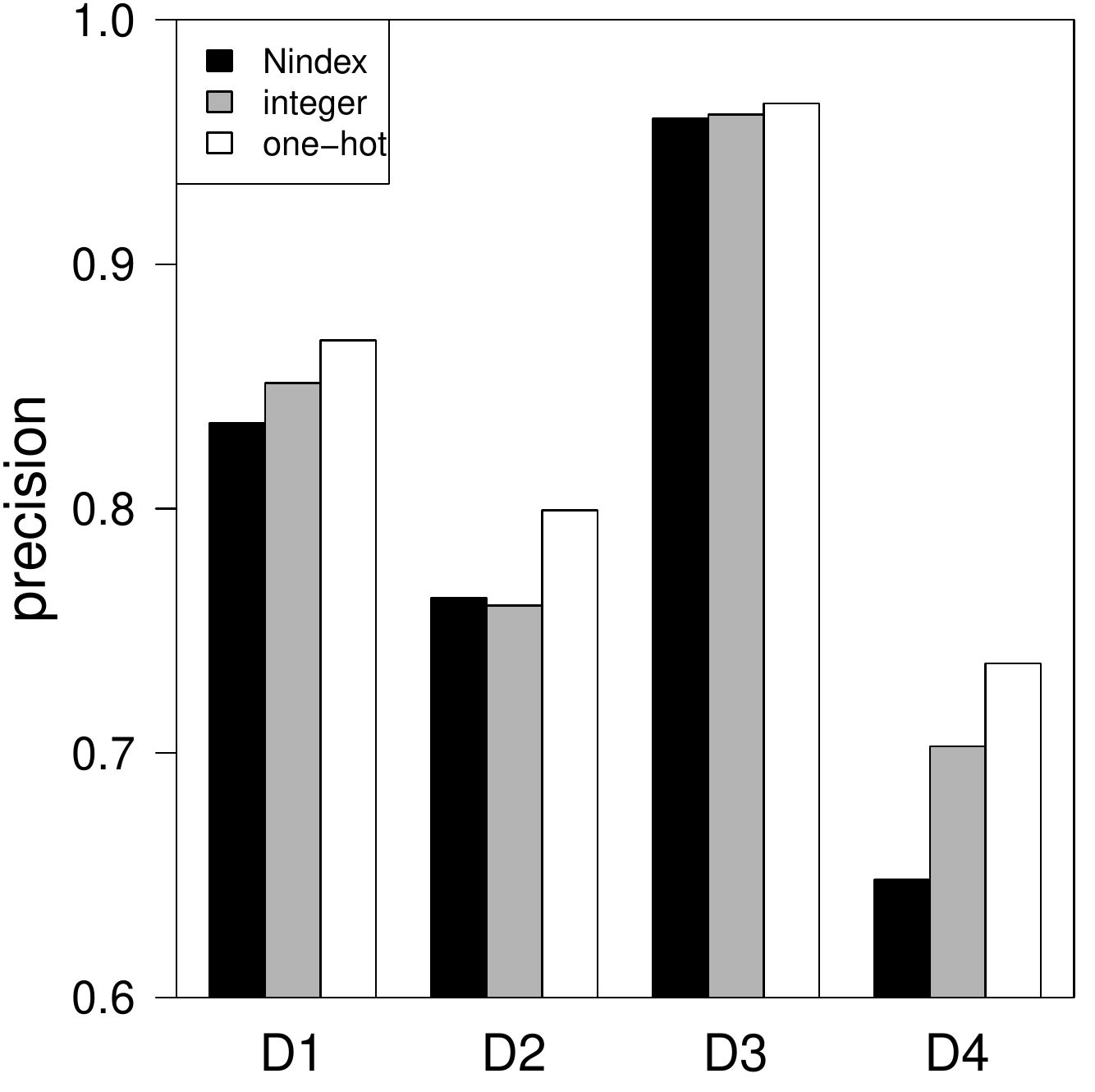}
}
\hspace{0in}
\subfigure[Recall]{
\includegraphics[width=4cm]{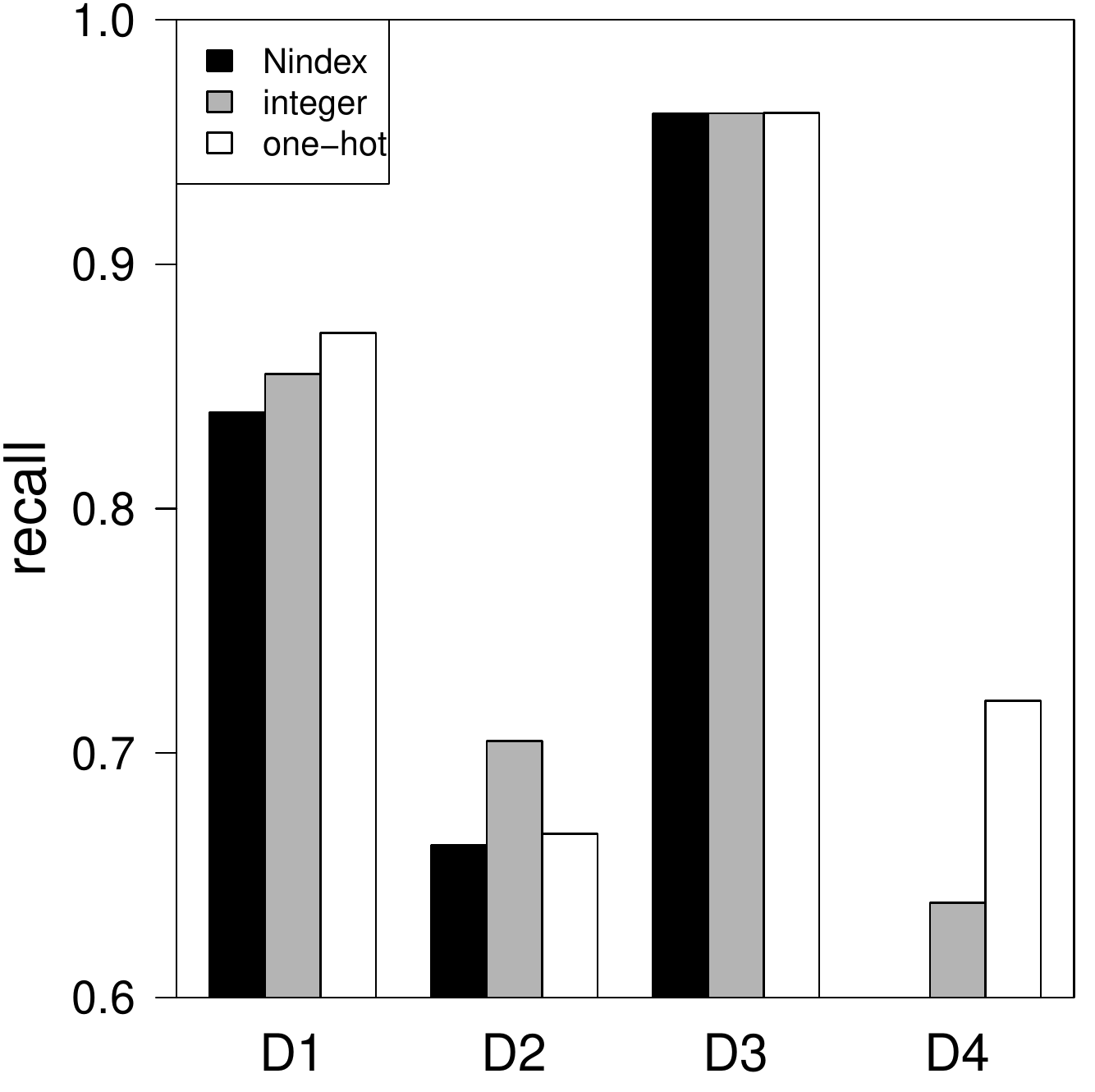}
}
\hspace{0in}
\subfigure[F-Measure]{
\includegraphics[width=4cm]{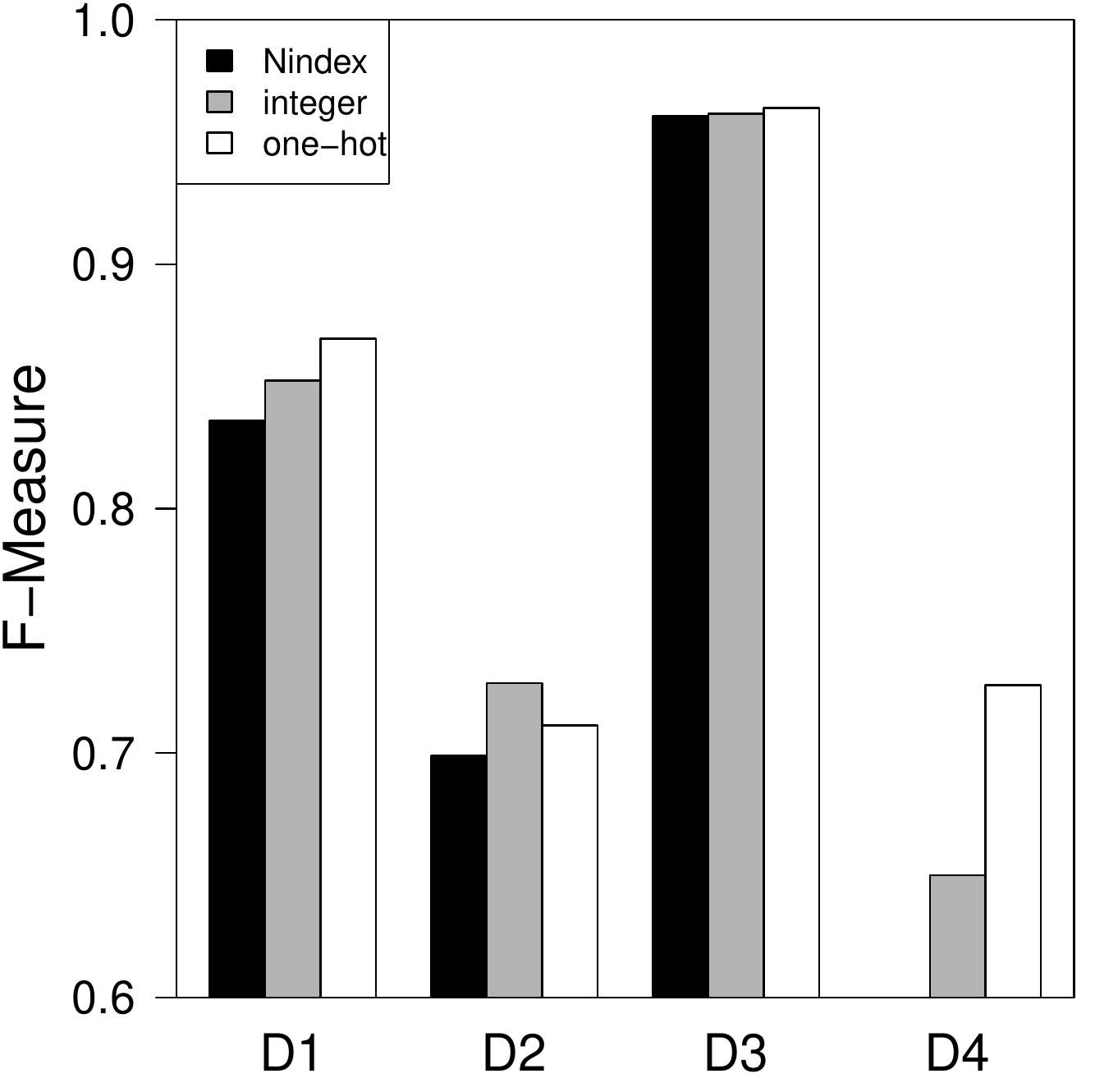}
}
}
\vspace{-2.5mm}
\caption{Study of second state representation.}
\label{fig6}
\vspace{-2mm}
\end{figure*}

Next, we study the impact of the second part of environment state, which is the index information of the current scanning feature. We consider three cases: i.e., (i) Nindex that doesn't consider the feature index in environment state space; (ii) Integer that incorporating integer encoding of the feature index into environment state space; (iii) One-hot that adds one-hot encoding of the feature index into environment state space.
From Fig.~\ref{fig6}, we see that although the one-hot encoding of feature index has the lower recall value and $F$-measure value than integer encoding, it can perform best for other three datasets. In other word, the one-hot encoding can describe the current scanning position very well.

\section{Related Work}
We illustrate the related work in terms of feature selection and reinforcement learning for feature selection.

\subsection{Feature selection}
Feature selection is an essential problem in the field of machine learning and data mining, aiming to select an optimal feature subset by removing irrelevant and redundant features. Traditional feature selection methods can be broadly divided into three categories: filter methods, wrapper methods and embedded methods. Filter methods rank the features based on some evaluation measures, e.g., fisher score \cite{b5}, ReliefF \cite{b26}, information gain \cite{b7}, Chi-square score \cite{b30}, and Pearson correlation coefficient \cite{b6}, and then select top-ranking features as an optimal feature subset. The typical filter methods are Maximum Relevance Minimum Redundancy (mRMR) \cite{b8}, fast correlation based filter algorithm (FCBF) \cite{yu2004efficient}, and univariate feature selection \cite{forman2003extensive}. Due to be independent of any learning algorithm, filter methods have low computational cost and are suitable for high-dimensional data. However, filter methods ignore interactions between features and the classification performance of selected features. Wrapper methods search feature space and employ a predetermined learning algorithm to evaluate the selected feature subset directly. Therefore, they are more accurate than filter methods. The typical wrapped methods include Recursive Feature Elimination (RFE) \cite{b9}, sequential Feature Selection \cite{b10, b31, b32}, branch-and-bound search \cite{b38} and evolutionary algorithms for feature selection \cite{b11, b12, b13, b14}. However, with the increasing of feature dimension, wrapper methods need to explore a feature space of $2^N$ feature subsets and are likely to suffer from local optima. Embedded methods simultaneously optimize classification performance and feature subset by integrating feature selection with classification model, e.g., LARS \cite{efron2004least}, LASSO \cite{tibshirani1996regression}, and decision tree \cite{sugumaran2007feature}. Although these methods make a tradeoff between the efficiency and the predictive accuracy, their selected features might not be suitable for other classifiers.

\subsection{Reinforcement learning for feature selection}
Model-free reinforcement learning was born to make long-term optimal decisions with no or little prior knowledge of the dynamic environment \cite{b37} \cite{wang2020incremental}. This can provide the outstanding capability in feature selection field. Some feature selection methods via reinforcement learning are proposed. Reference \cite{b17} \cite{b18} formulate the single agent to make decisions. The actions of this agent include the selection or deselection of all $n$ features. The $2^n$ size of action space is too large to globally explore the feature subspace. In other words, the action space of this agent is $2^n$. Such formulation is similar to the evolutionary algorithms \cite{b33}, which are likely to obtain local optima. \emph{Fang et al.} also developed a single-agent reinforcement learning for feature selection in malware detection, where the action space represents the feature index that is selected and a terminal action \cite{b34}. Although the action space is $D+1$, this formulation is similar to SFS, where a feature that is selected cannot be removed in later stages and is likely to obtain local optima. \emph{Liu et al.} proposed a multi-agent reinforcement learning framework (MARL) for feature selection problem by regarding each feature as an agent \cite{b19}. 
However, the complexity of the MARL increases as the number of agents, because each agent assigned to each feature adds its own variables. So, multi-agent demand high computer configuration for the high dimensional data. At the same time, the nonstationarity arises in MARL because of all the agents are learning simultaneously \cite{b35}.

\section{Conclusion and Future Work}
In this study, we present a novel deep reinforcement learning framework for feature selection. By modeling feature selection with the scanning scheme, we can not only limit the action space of the agent, but also solve feature selection problem with single-agent reinforcement learning framework. That allows the proposed reinforcement learning to work on a low configuration computer. In addition, we incorporate two important factors, i.e., feature relevance calculated by information gain and redundancy calculated by Pearson correlation coefficient to guide the automatic feature subspace exploration and improve the quality of selected features. In order to better represent the environment state, we design a convolutional auto-encoder with spatial pyramid pooling layer to handle the dynamic environment, which is the selected feature subset. Finally, comprehensive experiments on real-world datasets demonstrate the effectiveness of proposed algorithm.

\section*{Acknowledgment}
This research was partially supported by the National Science Foundation (NSF) via the grant numbers: 1755946, I2040950, 2006889.


\bibliographystyle{IEEEtran}
\vspace{12pt}

\end{document}